\documentclass{article}

\usepackage[square,numbers]{natbib}
\usepackage{xcolor}
\usepackage[linesnumbered,algoruled,boxed,ruled,vlined]{algorithm2e}
\usepackage{amsmath}
\DeclareMathOperator*{\argmax}{argmax}
\usepackage{graphicx}
\usepackage{caption}
\usepackage{subcaption}
\usepackage{float}
\usepackage{nccmath}

\usepackage{enumitem}
\usepackage{multirow}

\usepackage{caption} 
\usepackage{setspace}

\usepackage{pgf}
\usepackage{tikz}
\usepackage{collcell}
\definecolor{myGreen}{rgb}{0,0.5,0}

\newcommand*{\MinNumber}{1.0}%
\newcommand*{\MidNumber}{5.5} %
\newcommand*{\MaxNumber}{9}%

\newcommand{\ApplyGradient}[1]{%
    \ifdim #1 pt > \MidNumber pt
    \pgfmathsetmacro{\PercentColor}{max(min(100.0*(#1 - 
    \MidNumber)/(\MaxNumber-\MidNumber),100.0),0.00)} %
    \hspace{-0.33em}\colorbox{myGreen!\PercentColor!yellow}{#1}
    \else
    \pgfmathsetmacro{\PercentColor}{max(min(100.0*(\MidNumber - 
    #1)/(\MidNumber-\MinNumber),100.0),0.00)} %
    \hspace{-0.33em}\colorbox{red!\PercentColor!yellow}{#1}
    \fi
}

\newcolumntype{G}{>{\collectcell\ApplyGradient}c<{\endcollectcell}}
\usepackage{adjustbox}
\usepackage{array}

\newcolumntype{J}[2]{%
    >{\adjustbox{angle=#1,lap=\width-(#2)}\bgroup}%
    l%
    <{\egroup}%
}
\newcommand*\rotz{\multicolumn{1}{J{0}{-1em}}}

\restylefloat*{figure}
\restylefloat*{table}
\usepackage[english]{babel}

\usepackage{arxiv}
\usepackage[utf8]{inputenc} 
\usepackage[T1]{fontenc}    
\usepackage[colorlinks=true,linkcolor=black,anchorcolor=black,citecolor=black,filecolor=black,menucolor=black,runcolor=black,urlcolor=black]{hyperref}
\usepackage{url}            
\usepackage{booktabs}       
\usepackage{amsfonts}       
\usepackage{nicefrac}       
\usepackage{microtype}      

\usepackage{authblk} 
\usepackage{fancyhdr}
\title{Probabilistic Diagnostic Tests for Degradation Problems in Supervised Learning}

\author[a,b,*]{Gustavo A. Valencia-Zapata}
\author[c]{Carolina Gonzalez-Canas}
\author[d]{Michael G. Zentner}
\author[b]{Okan Ersoy}
\author[a,b]{Gerhard Klimeck}

\affil[a]{Network for Computational Nanotechnology, Purdue University, West Lafayette IN, US}
\affil[b]{School of Electrical and Computer Engineering, Purdue University, West Lafayette IN, US}
\affil[c]{School of Industrial Engineering, Purdue University, West Lafayette IN, US}
\affil[d]{HUBzero Research Group, Information Technology at Purdue, West Lafayette IN, US}

\affil[*]{Corresponding author: Gustavo A. Valencia-Zapata, info@gustavovalencia.com}

\begin{document}
\maketitle

\begin{abstract}
Several studies point out different causes of performance degradation in supervised machine learning. Problems such as class imbalance, overlapping, small-disjuncts, noisy labels, and sparseness limit accuracy in classification algorithms. Even though a number of approaches either in the form of a methodology or an algorithm try to minimize performance degradation, they have been isolated efforts with limited scope. Most of these approaches focus on remediation of one among many problems, with experimental results coming from few datasets and classification algorithms, insufficient measures of prediction power, and lack of statistical validation for testing the real benefit of the proposed approach. This paper consists of two main parts: In the first part, a novel probabilistic diagnostic model based on identifying signs and symptoms of each problem is presented. Thereby, early and correct diagnosis of these problems is to be achieved in order to select not only the most convenient remediation treatment but also unbiased performance metrics. Secondly, the behavior and performance of several supervised algorithms are studied when training sets have such problems. Therefore, prediction of success for treatments can be estimated across classifiers.
\end{abstract}



\section{Introduction}

A diagnosis is an investigation or analysis of causes or nature of a condition, situation, or problem. For example, in Health Care, the diagnostic process gathers information related to a patient health problem in order to plan the best path or treatment. Recent retrospective studies of adult healthcare and post-mortem examinations  show a diagnostic error leading to up to 17\% of hospital adverse events and approximately 10\% of patient deaths \cite{CommitteeonDiagnosticErrorinHealthCare2015}. Therefore, several efforts in Health Care focus on improving diagnosis such as reducing uncertainty, early diagnosis, and second opinion \cite{Bordini2017}, \cite{Graber2013}. Using the analogy of "Medical Diagnosis", it is plausible to argue that the concept of "Diagnostic" as human cognitive function can be incorporated in the Artificial Intelligence domain. This research investigates whether the implementation of diagnostic tests aids to characterize the most common problems of classification tasks in order to select the most convenient treatment for avoiding classification degradation. Thereby, the questions now arise: 
\vspace*{2mm}
\begin{itemize} \item What are the most representative problems of classification? Does exist any correlation, causality or association between them? \item Is it possible to measure a direct impact of each problem to the classifier degradation? \item Is there any process or technique to diagnose training datasets and subsequent remediation (treatments) selection before the training stage
\end{itemize}
\vspace*{3mm}

\newpage

Mainly there are 5 degradation problems related to training datasets that limit the learning process: class imbalance, sparseness, small-disjuncts, overlapping, and noise labels. The class imbalance is one of the most recurrent problems and has received big interest in the artificial intelligence field in recent years \cite{Douzas2018,Fernandez2018a,Lu2017,X.2015,Beyan2015,10.1007/978-3-540-76725-1_42,Khoshgoftaar2007,Liu2006, Prati2004b,NiteshV.Chawlaet.al2002,Chawla2004}. A dataset is imbalanced when its class distribution or classification categories do not have similar proportions in terms of the number of instances. This is a problem because classification algorithms are not designed to handle unbalanced datasets and tend to favor the major class over minor ones \cite{Chawla2004}. A second important issue related to class imbalance is the performance measurement. The most used performance metric for classification algorithms is the area under the \textit{ROC} curve (\textit{AUC}) \cite{Swets1988}, which may provide an overly optimistic performance evaluation and wrong performance evaluation for unbalance cases \cite{Davis2006}. Several questions have arisen about the class imbalance problem; for its proper functioning: what is the optimal class distribution for training a classification algorithm?  What are treatments for handling class imbalance? Is the class imbalance only responsible for classifier degradation?

Several approaches \cite{Japkowicz2000,Weiss2003,Estabrooks2004}, have tackled these questions applying resampling (oversampling and undersampling) to reach the balance between classes modifying the distribution of the dataset. More sophisticated techniques change that distribution by generating synthetic data as a balance strategy, not only for reducing overfitting and improving performance in classification tasks, but also for handling the sparseness in the feature space \cite{NiteshV.Chawlaet.al2002}. On the other hand, cost-sensitive learning treat the class imbalance problem arguing that each type of error has its own costs  \cite{Domingos1999,Elkan2001}. Therefore, instead of changing the distribution of the whole dataset, cost-sensitive based models assume that misclassification cost for the observed dataset is known and use a "cost matrix" for describing it. Different approaches suggest that class imbalance is not completely responsible for degradation and argue that the phenomena of small-disjuncts contributes to overfitting and misclassification \cite{Prati2004b,Holte1989,Japkowicz2001,Nickerson2001,Jo2004,article}. Small-disjuncts are subclasses of few instances inside the minor class. Other studies state that the "overlapping" regions between classes \cite{10.1007/978-3-540-76725-1_42,Prati2004a} and "noisy labels" due mislabeling \cite{Brodley1999} have strong negative effect over the classification algorithm performance. Different approaches based on ensemble methods \cite{Brodley1999,Lu2017,Liu2006,Opitz1999,Polikar2006,Rokach2010} combine multiple learning algorithms and strategies for handling class imbalance, noisy labels detection \cite{Brodley1999}, and improve performance.

\vspace*{3mm}

This research presents a comprehensive empirical study of degradation problems in supervised learning trough datasets from real-world domains. The paper is organized as follows: Section 2 describes the importance of diagnostic methods in machine learning, research aims and approach. Section 3 describes the five major degradation problems and remediation techniques for handling degradation in supervised learning algorithms. Section 4 presents the implementation of a probabilistic diagnostic model for classifier degradation, which supports the framework proposed in this research.  Section 5 presents experimental results for several datasets across classification algorithms and techniques for handling degradation problems. Finally, Section 6 presents general conclusions and recommendations for future work.

\section{Lack of diagnosis in supervised learning}

The motivation of the present work is to reduce degradation of classification algorithms caused by inherent problems associated with datasets. Although there exist methodologies or techniques that try to mitigate degradation, most of those efforts are blind to the characteristics of the datasets under study.

The diagnostic process in Health Care has several components \cite{CommitteeonDiagnosticErrorinHealthCare2015} such as stages related to information gathering, diagnostic testing, clinical history, etc. This research uses similar components applied to the artificial intelligence domain, specifically associated with supervised learning. In other words, the process starts with the "dataset" or target, which experiences a problem. After the dataset engages with the process, information is gathered from several sources (characteristics, testing, records, best practices, and so on) for building a diagnosis. Consecutively, an explanation of the problems is built. Finally, a treatment is defined based on the diagnosis, and its outcomes are available as inputs to the diagnostic process. Two fundamental principles are adopted from Health Care:
\vspace*{3mm}
\begin{enumerate}
\itemsep=5pt
\item Correct diagnosis avoids inappropriate treatments \cite{CommitteeonDiagnosticErrorinHealthCare2015}. Even though a number of approaches (treatments) either in the form of a methodology or a technique try to minimize performance degradation in supervised learning, they have been isolated efforts without the input from a diagnostic process.
\item Treatment procedures according to the needs and characteristics of individual patients \cite{Beutler2016,LarryJameson2015}. Treatment effectiveness for classifier degradation does not exclusively depend on the degradation problem (class imbalance, sparseness, small-disjuncts, overlapping, and noise labels), but also on characteristics associated with the training set relevant for treatment selection and reaching optimal results.
\end{enumerate}  
\vspace*{3mm}

\section{Degradation problems for supervised learning}

Data complexity can be associated with issues such as class imbalance, overlapping, sparseness, and others \cite{Fernandez2018}. This section presents five degradation problems for supervised learning and their most well-known remediation techniques. The damage caused by those degradation problems can range from hampering the training stage to dramatically reduce classification algorithm performance.

\subsection{Class imbalance and Sparseness}
Imbalanced data is present in many fields and represents a continuous challenge for Data Mining and Artificial Intelligence tasks. For instance, in medical diagnosis, some diseases are infrequent or hard to detect because of limited number of cases. Fraud detection is a well-known case of class imbalance, in which the number of fraud transactions is much smaller than the legitimate ones. As previously mentioned, part of misclassification in supervised learning tasks may be attributed to the effect of class imbalance and the fact that most of the classifiers are not designed to address this effect. Class imbalance occurs when one or more classes from the training set have few instances as compared to other classes. Usually, the level of class imbalance is measured using the Imbalance Ratio (IR), which indicates the relationship between two classes that expresses how much bigger one is than the other. For example, in a binary classification task (dataset with two classes), the class with few observations or minority class is called the "Positive class". On the other hand, the majority class is called the "Negative class" and the IR is estimated using the ratio between the total numbers of observations for each class. Generally, the class imbalance problem has been addressed with two main approaches: methods based on resampling (oversampling and undersampling) and synthetic data techniques, and cost-sensitive methods. In the case of the latter approach, it argues that errors do not have the same cost. The present research focuses on resampling and synthetic data techniques, because cost-sensitive techniques assume the prior knowledge about the cost of misclassification for each class, which is a condition unknown in most cases and defined by the nature and domain of the dataset. The most representative resampling methods are oversampling and undersampling \cite{Breiman1984,Kubat1997}. These two techniques seek to reach the balance between classes modifying the distribution of the dataset. The first replicates randomly some instances from the Positive class. The main disadvantage of oversampling is overfitting \cite{NiteshV.Chawlaet.al2002,Kubat1997} and the low performance for unseen testing instances at the training stage \cite{Holte1989}. In contrast, undersampling subtracts a random sample of a specified size from the observations of the Negative class, leading losing important information and limiting the learning process for such as class. C. Drummond et al. \cite{Drummond2003} tested oversampling and undersampling techniques over class imbalance datasets using the decision tree learner C4.5, showing better performance for classification tasks when undersampling technique was selected. However, more modern strategies such as Synthetic Minority Oversampling Technique (SMOTE) \cite{NiteshV.Chawlaet.al2002}, Borderline-SMOTE \cite{Chawla2004}, Adaptative Synthetic Sampling Approach for Imbalanced Learning (ADASYN) \cite{He2008},  Density-base SMOTE (DBSMOTE) \cite{Bunkhumpornpat2012}, and BalanceCascade and EasyEnsemble \cite{Liu2006} combine oversampling, undersampling, and algorithms in order to produce synthetic entries, reach balance, avoid overfitting, and obtain better performance.
\vspace*{2mm}

Several studies argue that the imbalance issue in data should be understood as the level of sparseness and the low degree of instances for some classes \cite{NiteshV.Chawlaet.al2002,Chawla2004,Japkowicz2002}. Therefore, synthetic data techniques such as SMOTE \cite{NiteshV.Chawlaet.al2002} and its extensions not only generate new synthetic entries for balancing, but also builds larger and better defined decision regions for the Positive class by filling holes in the space.

\subsection{Small-disjuncts}

Holte et al. \cite{Holte1989} introduced the term of small-disjuncts, which would be the major cause of learning problems in the Positive class \cite{Holte1989,Weiss:2000:QSS:647288.721597}. The small-disjuncts could be interpreted as a direct consequence of the "Sparseness". In simple words, small-disjuncts are "sub-clusters" or subclasses of few instances inside the Positive class. Generally, the small-disjunct concept can be generalized as the imbalance problem between-class and within-class, the between-class case refering to the imbalance between the Negative class and the Positive class, which is usually estimated by the IR. On the other hand, the within-class refers to how observations are distributed within each class, since a class could be composed of several subclasses with imbalance in the number of instances \cite{Japkowicz2001,Nickerson2001,Jo2004,article}. This research extends the concept of small-disjuncts to the Negative class. Therefore, using the new definition of small-disjuncts: both classes can have subclasses of few instances that are difficult to handle during  the training stage for classification algorithms.

\subsection{Overlapping and noisy labels}

Datasets can have ambiguous regions, which contain observations from two or more classes with similar probability. This problem is known as overlapping. Diverse studies claim that classifier degradation is not only affected by the class imbalance problem or small-disjuncts, but perhaps even more so, in terms of the degree of overlapping \cite{10.1007/978-3-540-76725-1_42,Prati2004a,Batista2004}. Garcia et al. \cite{10.1007/978-3-540-76725-1_42} conducted several experiments over two-dimensional artificial datasets with two classes separated by a line orthogonal to one of the axis and six different classifiers. They found that changing the IR of the overlapping region is more beneficial than changing the overall IR in the dataset. Although the simplicity of the data is far from real datasets, the results point out the degree of overlap as an important factor in the process of building classifiers. Most of the treatments for handling Noisy Labels focus on cleaning the overlap regions by using the k-Nearest Neighbor Rule as a component in their procedure. Labeling error is a recurrent problem in supervised learning datasets. This problem describes instances with wrong label assignment due to different causes (i.e. data-entry error, subjectivity, and lack of information for the labeling process). Quinlan \cite{J.R.Quinlan1986,Quinlan1983,q-encl-86} conducted several experiments in order to measure the effect of class noise over decision tree performance, finding two patterns: class noise is more harmful than feature (characteristics) noise, and cleaning noisy observations from training sets produces classifiers with better accuracy. Condensed Nearest Neighbors algorithm (CNN) \cite{Hart:1968:CNN} was the first attempt of filtering noisy observations or building a better subset from the training set using Nearest Neighbor Rule (NN Rule) \cite{Cover1967,Cover1968}. Several subsequent proposals based on CNN concepts have been discussed, such as Edited Nearest Neighbors (ENN) \cite{Wilson1972}, Neighborhood Cleaning Rule NCL \cite{Laurikkala:2001:IID:648155.757340},  Tomek \cite{1976AEwt}, Mutual Nearest Neighborhood  (MNN) \cite{Gowda1979}, and One-Side Selection (OSS) \cite{Kubat1997}. For instance, Tomek algorithm extends that approach calling ENN for increasing the number of neighbors and focus on removing noisy observations from the Negative class close to the borderline with the Positive class and returning a better subset for the training stage. On the other hand, Brodley et al. \cite{Brodley1996} used Ensemble filters for detecting noisy observations. In this approach,  the filter is composed of a set of classifiers called base-level detectors, which receive the training set and detect the misclassified cases. Finally, the filter removes observations based on majority voting or consensus and generate a training subset.

\subsection{Treatments and side effects}

As previously stated, there are numerous treatments in the form of methodology or technique that seeks to mitigate negative effects of the degradation problems. However, the effectiveness of such treatments may be reduced by lack of diagnosis. The following is the list of techniques implemented in Section 5 related to experimental framework.

\vspace*{2mm}
\begin{enumerate}[label=\arabic*)]
\itemsep=5pt
\item \textit{Resampling} (Random). This technique combines oversampling and undersampling to obtain a balanced dataset. Naive oversampling replicates randomly some instances from the Positive class (minority). On the other hand, naive undersampling removes randomly some instances from the Negative class (majority).

\item \textit{Synthetic Minority Oversampling Technique} (SMOTE) \cite{NiteshV.Chawlaet.al2002}. This algorithm generates synthetic instances of the Positive class (minority) based on the feature space similarities, using the k-Nearest Neighbor algorithm. Besides, the Negative class (majority) instances are under-sampled in order to reach the balance between classes.

\item \textit{Borderline-SMOTE} \cite{Chawla2004}. This technique is an extension of SMOTE, which focuses on generating synthetic Positive instances in the neighborhood of the borders.

\item \textit{Adaptative Synthetic Sampling Approach for Imbalanced Learning} (ADASYN)\cite{He2008}. This algorithm is an extension of SMOTE, which generates synthetic instances using a density distribution based on k-Nearest Neighbors algorithm. Thus, this technique automatically decides the number of synthetic instances that must be generated for each real Positive instance.

\item \textit{Density-base SMOTE} (DBSMOTE) \cite{Bunkhumpornpat2012}. This technique initially clusters the Positive class using DBSCAN algorithm \cite{EsterM.KriegelH.P.SanderJ.&Xu1996} and generates synthetic instances along a shortest path from each Positive instance to a pseudo-centroid of the cluster.

\item \textit{Edited Nearest Neighbors} (ENN) \cite{Wilson1972}. It removes any instance whose class label differs from the class of at least half of its k-Nearest Neighbors. Therefore, both classes may be under-sampled.

\item \textit{Neighborhood Cleaning Rule} (NCL) \cite{Laurikkala:2001:IID:648155.757340}. This algorithm is a modification of ENN, which only excludes instances from the Negative class (majority) in two steps: first, it removes Negatives instances which are misclassified by their 3-Nearest Neighbors. Then, the neighbors of each Positive instance are found and the ones belonging to the Negative class are removed.

\item \textit{One-Side Selection} (OSS) \cite{Kubat1997}. This technique combines Tomek links \cite{1976AEwt} and CNN \cite{Hart:1968:CNN}. First, Tomek links focus on cleaning overlapping between classes by removing Negative instances using 1-Nearest Neighbor. Then, CNN removes instances from both classes that are not correctly classified by the 1-Nearest Neighbor Rule.

\end{enumerate}
\vspace*{2mm}

This research categorizes the above treatments based on the degradation problems in order to remedy and possible side effects. Table \protect\ref{tbl1} summarizes such categorization. The symbol (+) refers to the problem that the treatment intends to mitigate. Contrarily, the symbol ($-$) refers to negative side effects due to the treatment. For instance, Random combines undersampling and oversampling to obtain a balanced dataset; however, the first one removes instances that may be important and the second one produces overfitting. A second very concrete example is OSS, which solves the problems of overlapping and noisy labels by removing instances. However, it reduces the amount of data, produces holes in the space, and may increase the class imbalance. On the contrary, SMOTE and its extensions not only regenerate synthetic instances for reaching balance between classes, but also fill the space of the Positive class. Therefore, those techniques reduce the sparseness effect and build better decision regions.


\begin{table}[t]
\small
\caption{Treatments: Theoretical benefits and side effects. The symbol (+) refers to the problem that the treatment intends to remediate.
Contrarily, the symbol ($-$) refers to negative side effects due to the treatment.
 }\label{tbl1}
 \centering
 \begin{tabular}[t]{lccccccc}
    \toprule
               & Class     & Sparseness & Small      & Overlapping & Noisy  & Amount  & Overfitting  \\
Treatment      & Imbalance &            & disjuncts  &             & labels & of data &              \\          
    \midrule
Random         & (+)     &         &      &        &      & ($-$)   & ($-$)  \\
SMOTE          & (+)     & (+)     &      & ($-$)  & ($-$)  & ($-$)   &    \\
B-SMOTE        & (+)     &         &      &        &      & ($-$)   &  \\
DBSMOTE        & (+)     & (+)     & (+)  &        &      & ($-$)   &   \\
ADASYN         & (+)     & (+)     &      &        &      & ($-$)   &   \\
ENN            & ($-$)   & ($-$)   &      & (+)    & (+)  & ($-$)   &   \\
NCL            & (+)     & ($-$)   &      & (+)    & (+)  & ($-$)   &    \\
OSS            & ($-$)   & ($-$)   &      & (+)    & (+)  & ($-$)   &     \\
    \bottomrule
    \end{tabular}
\end{table}


\section{Probabilistic diagnostic tests}
After the main degradation problems and their remediation treatments have been mentioned, this section presents a model called "Probabilistic Diagnostic Model for Handling Classifier Degradation", which allows to determine the level of degradation problems present in classification datasets. References of previous dataset diagnosis efforts point to basic description of the dataset such as missing values, duplicate instances, and feature statistics such as frequency, percentage, mean, and standard deviation. Nevertheless, there is no explicitly reference related to tools, techniques, methodologies or any other approach for diagnosis of degradation problems on supervised learning.

\subsection{Diagnostic components}

\begin{figure}[t]
	\centering
		\includegraphics[scale=.7]{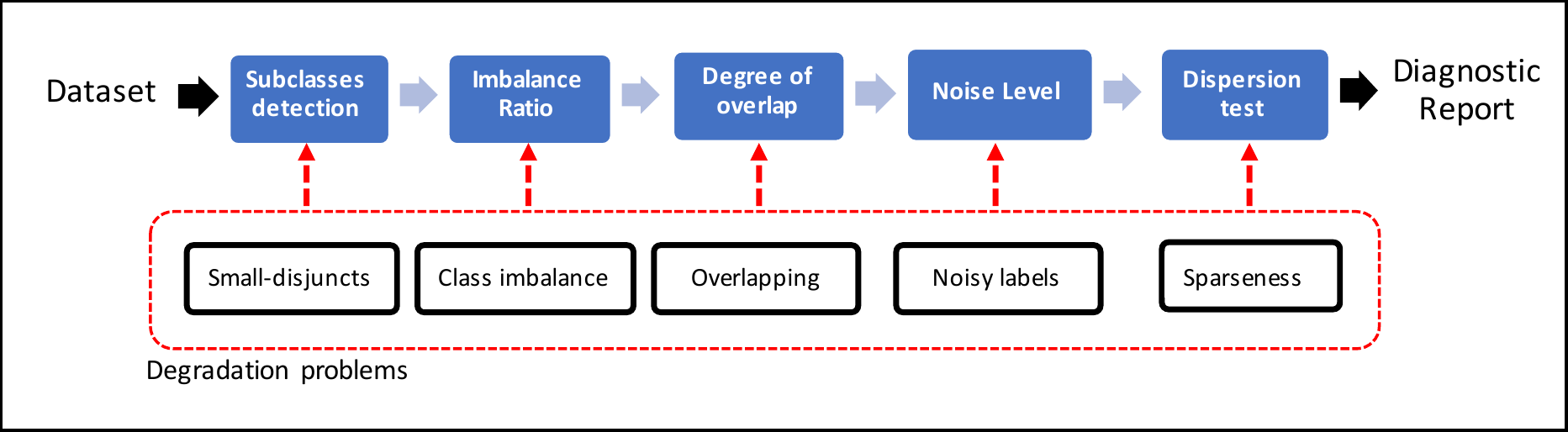}
	\caption{Diagnostic model process. The training set is the input to the flow and the diagnostic report is the output.}
	\label{FIG:1}
\end{figure}

The diagnostic model consists of five main components, which have a direct connection with the five degradation problems discussed in Section 3. These components are connected through a workflow where each component provides information to the next one during the diagnostic process. Figure \protect\ref{FIG:1} illustrates the sequence of the diagnostic model, where dataset (training set) is the input to the flow and the diagnostic report is the output. Subclasses detection is the first component, which is conducted for each class. Then, the process runs several "diagnostic tests" that measure the criticality levels of the degradation problems. Finally, the output of the process is a diagnostic report. 

\vspace*{2mm}
\begin{enumerate}
\itemsep=5pt

\item \textbf{Subclasses detection}: Previous proposals have formulated strategies of segmentation for the training set \cite{Douzas2018,Nickerson2001,Jo2004,article,Japkowicz2001,Ai2017} by using algorithms such as Principal Direction Devise Partition (PDDP) \cite{Boley1998a} and K-means \cite{MacQueen1967,Lloyd1982,EsterM.KriegelH.P.SanderJ.&Xu1996}. Those strategies focus on finding subclasses (clusters) for the positive class or even for the complete training set regardless of the membership class. Nevertheless, the major drawbacks of those approaches are the lack of parameters associated with the subclasses description and need to provide the number of subclasses (clusters) in the training set a priori.
The first component of the diagnostic model runs a segmentation process of partitioning each class into subclasses, implementing Gaussian Mixture Models (GMM) \cite{Pearson2006,titterington_85} and the EM algorithm \cite{A.P.DempsterN.M.Laird1977}, which is used to estimate the mixture parameters. The advantage of this approach is to obtain statistical parameters that describe each subclass for subsequent analysis. However, there is an inevitable, indeed necessary, question at the time of modeling a mixture of probabilistic distributions or any segmentation technique: how many clusters (subclasses) should be included in the model? A second question is related to the covariance matrix model selection for the GMM. For this research, the total number of subclasses per class and the covariance matrix model are determined by seeking to balance the increase in likelihood and the complexity of the mixture model, introducing a penalty term for each parameter. Then, the selection of the best model is addressed by the Bayesian Information Criterion (BIC) \cite{Schwarz1978,Fraley1998a}, which is described with more details in Appendix A.
Once the number of subclasses per class is defined, it is possible not only moving to the next component of the diagnostic tool, but also knowing more detail about the complexity of the dataset in terms of small-disjuncts. Greater number of subclasses is associated with higher difficulty levels of classification \cite{Japkowicz2002,article}. Moreover, it should be taken into consideration that not all detected subclasses are small-disjuncts; such category usually describes subclasses with few instances.

\newpage

\item \textbf{Imbalance ratio} (IR):  This component not only estimates the IR between classes, but also between subclasses. This information enables to focus any resampling or synthetic data process over specific subclasses in future remediation treatments. The IR is estimated as many times as the combination of detected subclasses taken 2 at a time without repetition.

\item \textbf{Degree of overlap}: the third component measures the degree of overlap based on the separation index (J*) proposed by Qiu et al. \cite{Qiu2006,Qiu2006a}. This index measures the magnitude of the gap between pairs of subclasses. It has a value between $-$1 and +1, where negative values indicate subclasses overlap, zero means subclasses are touching, and positive values indicate subclasses are separated. Experiments based on Multivariate Gaussian distributions (MGD) \cite{EsterM.KriegelH.P.SanderJ.&Xu1996} suggest that values larger than 0,21 indicate subclasses are well-separated. The separation index is estimated as many times as the combination of subclasses taken 2 at a time without repetition. Further information about estimation process of the separation index estimation can be found in Appendix A. One recurrent challenge of classification algorithms is to build a mathematical function sufficiently flexible and able to discriminate instances even with high levels of overlapping between classes. Therefore, some techniques focus on solving overlapping issues before the training stage, for instance, generating Positive instances in the overlap regions by reinforcing borderlines \cite{Chawla2004} or removing noisy instances from the Negative class \cite{Laurikkala:2001:IID:648155.757340,1976AEwt,Kubat1997}. The feasibility, advisability, and parametrization of those techniques could be elements easy influenced by the separation index.

\item \textbf{Noise level}: this component detects noisy instances using a 3-Nearest Neighbors Rule and estimates the noise ratio for the training dataset, the Positive class, and the Negative class. Since separation indices between subclasses are known, this component is able to identity noisy instances between well-separated subclasses. In other words, this component detects Noisy Labels unrelated to the overlapping issue, and estimates a ratio related to the noise due overlap issues.

\item \textbf{Dispersion test}: the fifth component implements two procedures: the first one estimates the degree of sparseness for each subclass by calculating the average and standard deviation of the distances (Euclidian, Mahalanobis, and Manhattan) to the subclass median. The second procedure implements the Anderson's test \cite{Anderson2006,Anderson2006a,Anderson2004} (PERMDISP2) for multivariate homogeneity of groups dispersion, which is a multivariate analogue of Levene's test for homogeneity of variances \cite{Brown1974}. Anderson's test seeks to validate if the dispersion of two or more subclasses is equal (similar sparseness).

\end{enumerate} 
\vspace*{2mm}


\subsection{Case in point: Artificial Domain}

This subsection presents a systematic example of the procedures and outputs of the diagnostic model. The purpose of this example is to provide a full understanding of the diagnostic model components and outputs. The case in point considers a two-dimensional artificial dataset for classification, which has been generated on the basis of Gaussian distributions and three controlled parameters: size of the subclasses, degree of overlapping, and distributional parameters (mean vector and covariance matrix).

\vspace*{2mm}
\begin{enumerate}
\itemsep=5pt

\item \textbf{Output 1: Training set basic description}. The first output is related to the basic characteristic of the dataset under analysis. In the supervised learning case, such dataset is the training set. Table \protect\ref{tbl2} shows the first output, which is a list of basic characteristics and values for the artificial training set. The Negative class with 404 instances represents 82.4\% of the data. The Positive class with 86 instances represents 17.6\% of the data. Therefore, this training set is said to have an Imbalance Ratio of 4.7:1 or IR= 4.7. In other words, there is one instance in the Positive class per 4.7 instances in the Negative class.


\begin{table}[t]
\small
\begin{minipage}[t]{.45\textwidth}
   \centering
   \caption{Output 1: Basic description for the artificial training set.}\label{tbl2}
   \begin{tabular}[t]{ll}
     \toprule
     Dimensions & 2 \\
     Duplicate Instances & NA \\
     Total instances & 490\\
     Instances Positive class & 86  \\
     Instances Negative class & 404 \\
     IR & 4.7 \\
     \bottomrule
   \end{tabular}
   
\end{minipage}\hfill
\begin{minipage}[t]{.45\textwidth}
   \centering
   \caption{Output 2: Subclass detection for the artificial training set.
}\label{tbl3}
   \begin{tabular}[t]{ll}
     \toprule
    \textbf{Positive class} &  \\
    Instances & 86 \\
    Covariance model & VEE\\
    Num. Subclasses & 2  \\ 
    \textbf{Negative class} &  \\
    Instances & 404 \\
    Covariance model & VEI\\
    Num. Subclasses & 3 \\
     \bottomrule
   \end{tabular}
\end{minipage}
\vspace*{3mm}
\end{table}



\begin{figure}
	\centering
		\includegraphics[scale=.6]{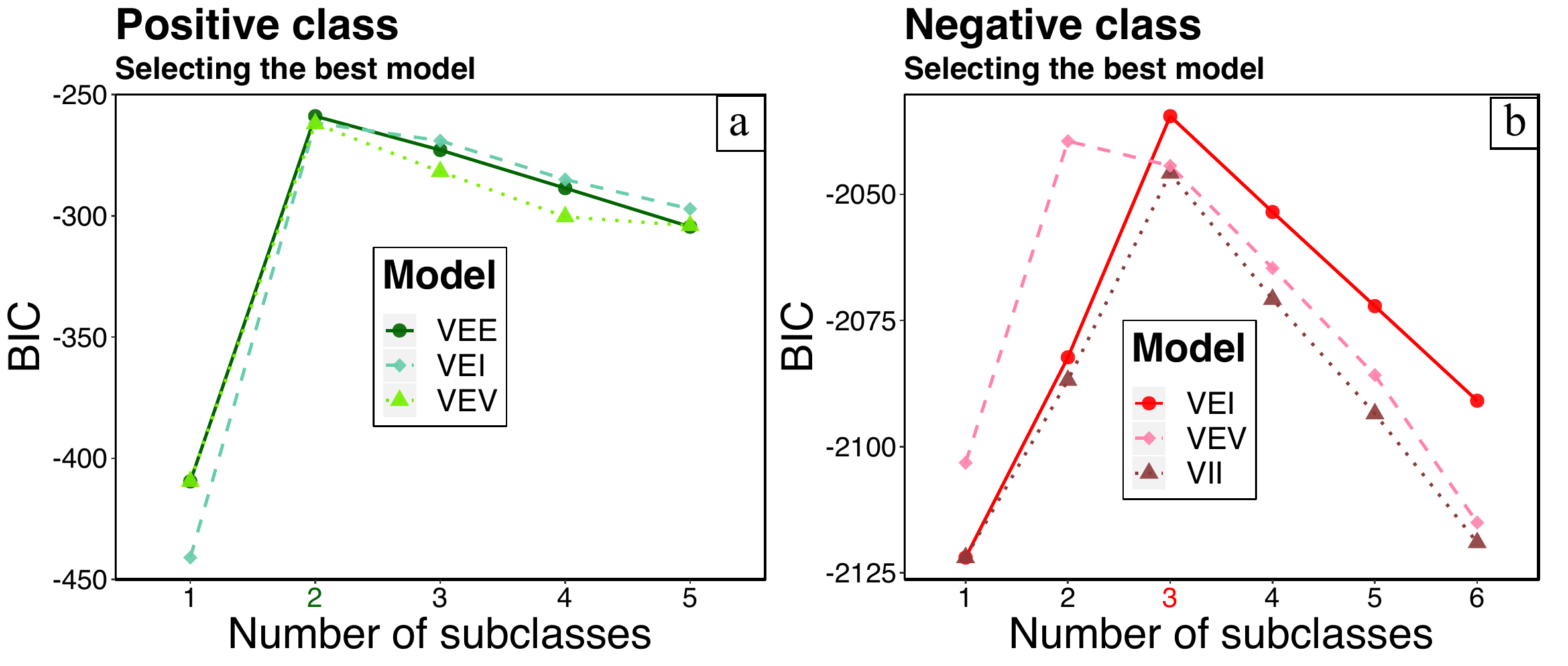}
	\caption{BIC plots for covariance matrix selection fitted to the artificial training set. a) The top three best models for the Positive class are: VEE (ellipsoidal, equal shape and orientation), VEI (diagonal, varying volume, equal shape), and VEV (ellipsoidal, equal shape). There are two subclasses for the Positive class. b) The top three best models for the Negative class are: VEI (diagonal, varying volume, equal shape), VEV (ellipsoidal, equal shape), and VII (spherical, unequal volume). There are three subclasses for the Negative class}
	\label{FIG:2}
\end{figure}

\item \textbf{Output 2: subclasses detection}. Table \protect\ref{tbl3} shows the next output related to the fist component of the diagnostic tool, which uses GMM for sub-subclass detection. For the Positive class, two subclasses were detected using the covariance matrix model VEE (ellipsoidal, equal shape and orientation). On the other hand, the Negative class has three subclasses detected by the covariance matrix model VEI (diagonal, varying volume, equal shape). Figure \protect\ref{FIG:2} illustrates the best number of subclasses and top three covariance matrix models based on the BIC for the artificial dataset. In Figure \protect\ref{FIG:2} (a), all covariance matrix models are in agreement with two subclasses for the Positive class, where the best covariance model is VEE. In Figure \protect\ref{FIG:2} (b), two out of three covariance matrix models are in agreement with three subclasses for the Negative class, and VEI is the best covariance model. Figure \protect\ref{FIG:3} makes a visual representation of the segmentation results through the GMM over the artificial dataset. The overall result is five subclasses, two for the Positive class and three for the Negative class. Further information about covariance matrix models implemented in GMM are given in Appendix A. 

\item \textbf{Output 3: IR and Overlap matrix} (IRO). Figure \protect\ref{FIG:4} shows the third output related to the IR and Overlap (IRO) matrix for the artificial training set. This is a square matrix, whose size is determined by the number of subclasses. The upper triangular part shows the IR between subclasses. For instance, the IR between \textbf{N-01} (Negative subclass 01) and \textbf{N-03} (Negative subclass 03) is 3. That means there is one instance in \textbf{N-03} per 3 instances in \textbf{N-01}. The diagonal of the matrix contains the number of instances per subclass. For example, P-01 (Positive subclass 01) is the smaller subclass with 36 instances in the artificial training set. Finally, the lower triangular part of the matrix shows the separation index J* between subclasses. For instance, J* = 0.29 between \textbf{N-02} and \textbf{P-02}, which makes sense since both subclasses are far from each other according to Figure \protect\ref{FIG:3}.

\newpage


\begin{figure}[t]
	\centering
		\includegraphics[scale=.45]{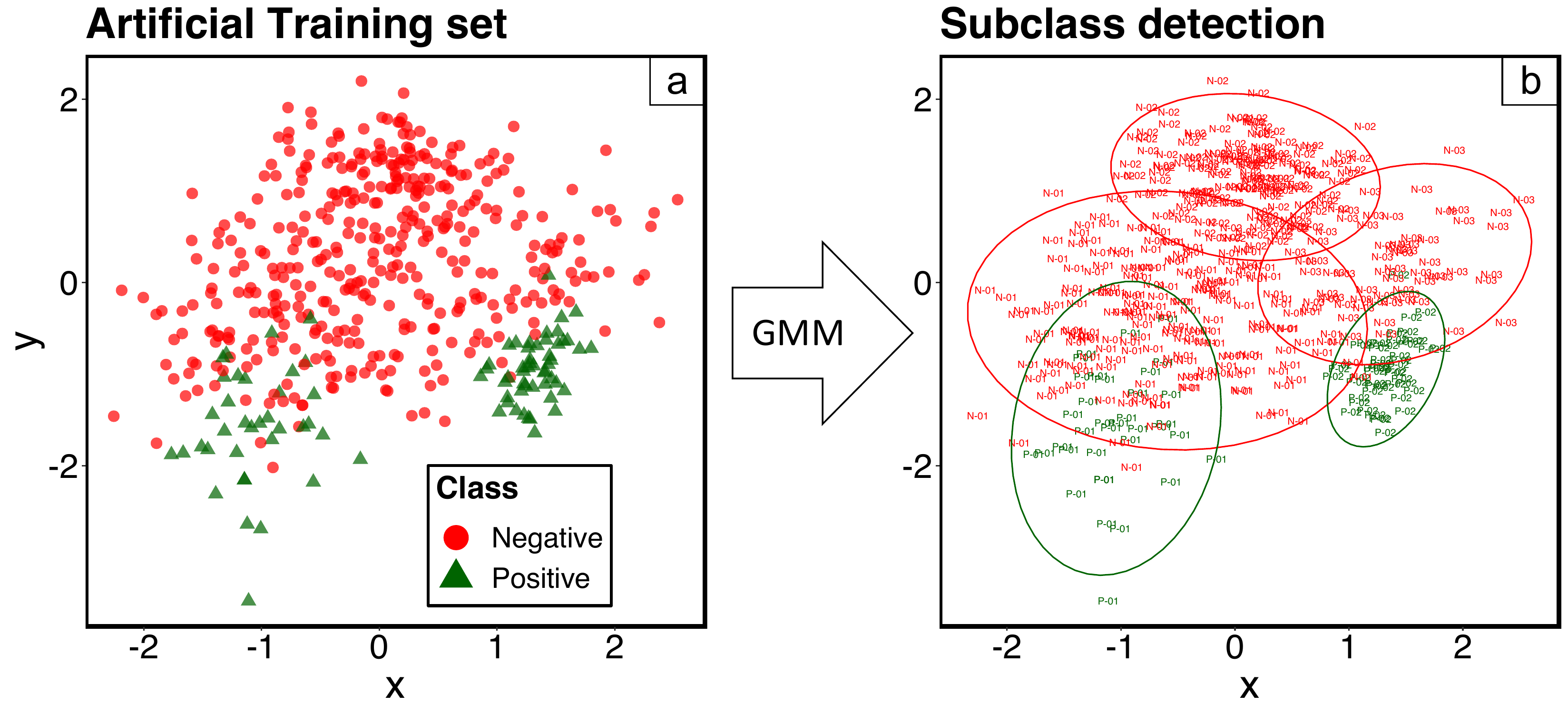}
	\caption{Detecting subclasses by using GMM. a) The original training set. b) Training set with five subclasses. \newline N = Negative subclasses and P = Positive subclasses.}
	\label{FIG:3}
\end{figure}


\begin{figure}
\begin{minipage}[t]{.5\textwidth}
   \centering
   	\includegraphics[scale=.32]{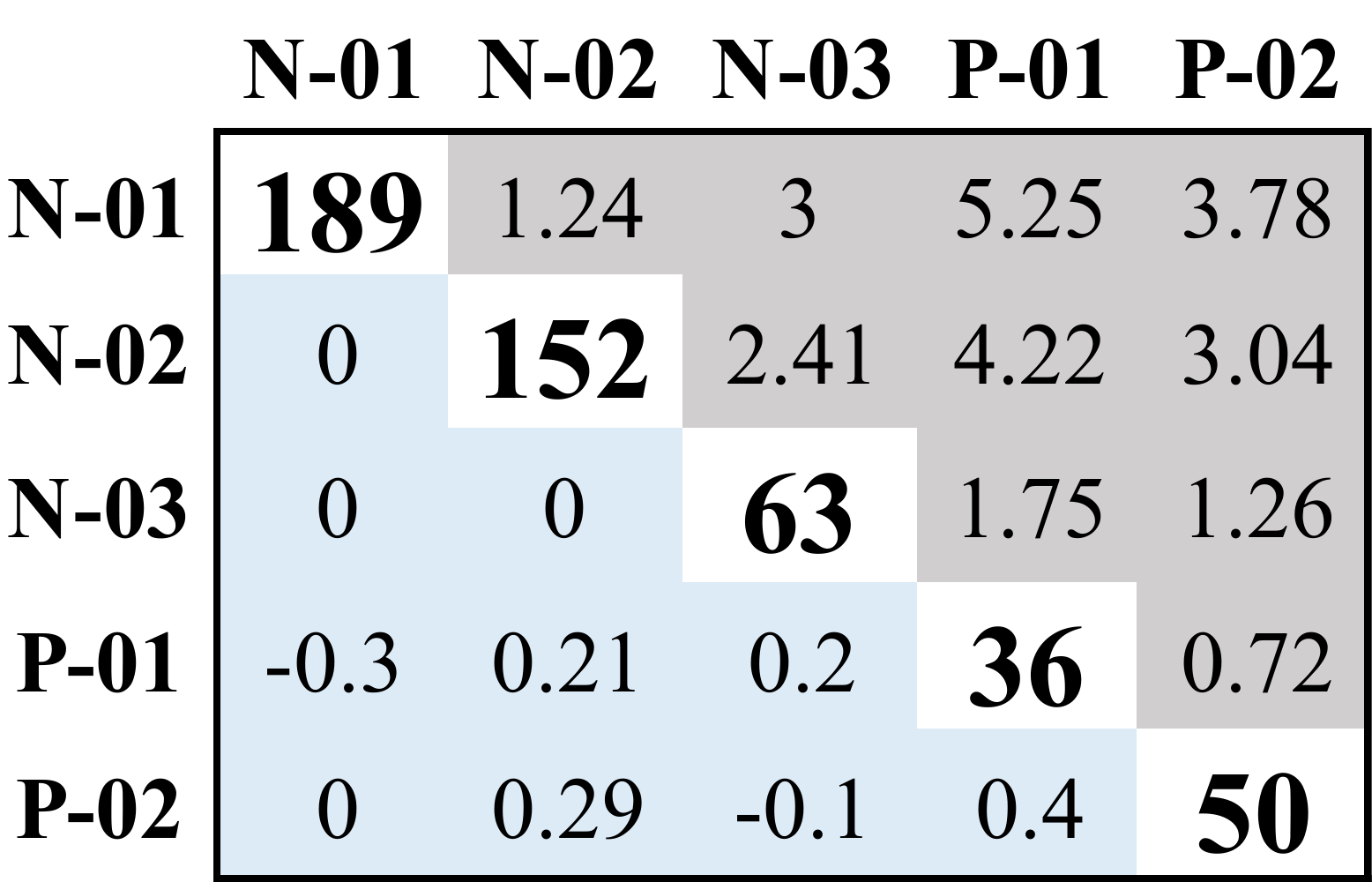}
	\caption{Output 3: IRO matrix for the artificial training set. The upper triangular part shows the IR between subclasses, the diagonal contains the number of instances per subclass, and the lower triangular part shows separation indexes between subclasses.}
	\label{FIG:4}
\end{minipage}\hfill
\begin{minipage}[t]{.4\textwidth}
   \centering
   \includegraphics[scale=.35]{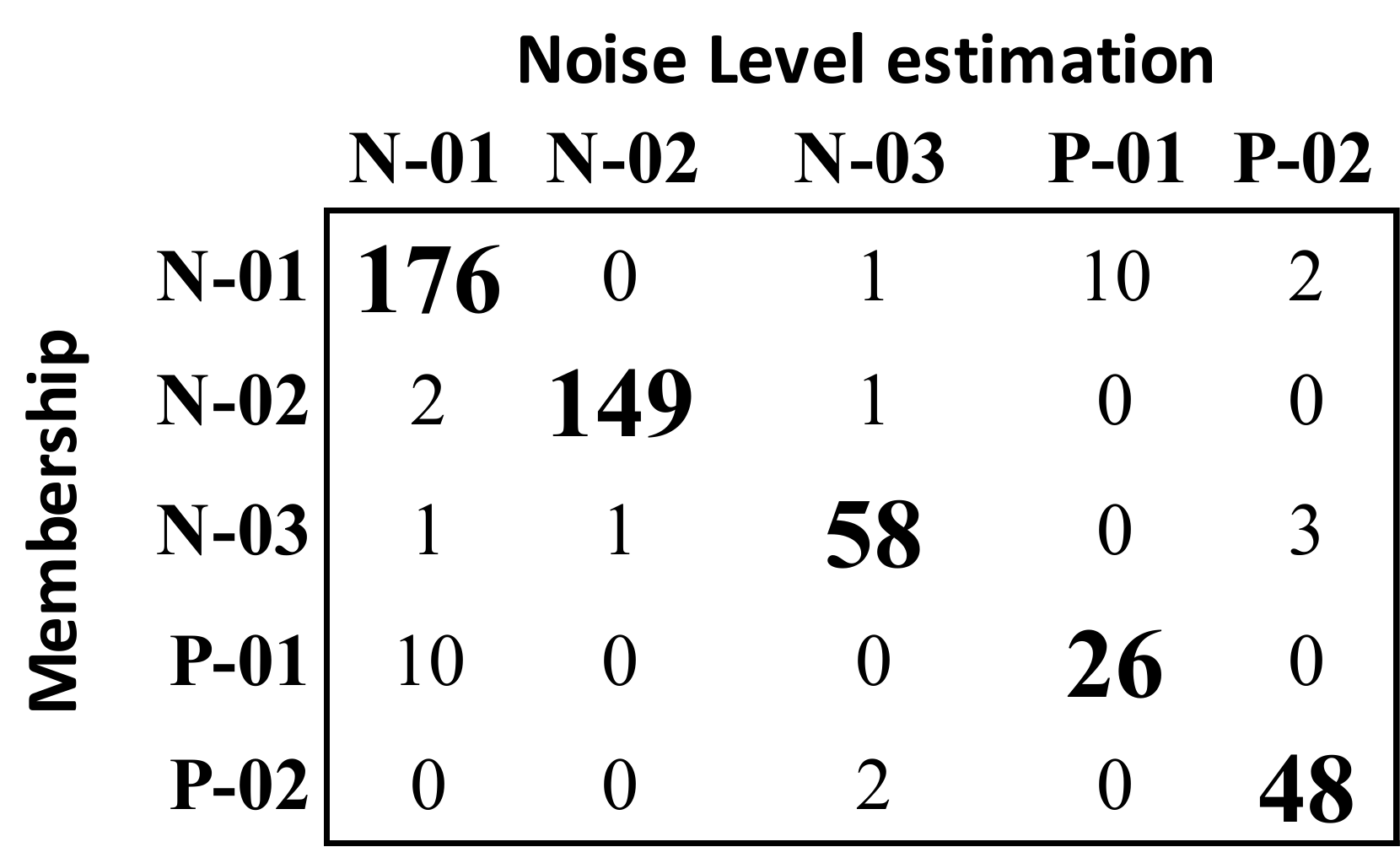}
	\caption{Output 4: Noise matrix for the artificial training set. This matrix shows the amount of noise between subclasses.}
	\label{FIG:5}
\end{minipage}
\end{figure}


\item \textbf{Output 4: Noise level}. The noise component has several tables as outputs. Table \protect\ref{tbl4} summarizes the number of noisy and valid instances per class for the artificial training set. Additionally, Table \protect\ref{tbl5} breakdowns the number of noisy instances per subclass and includes the noise location (last two columns). The column called "Noise overlap" refers to noise between subclasses with a separation index less than 0.2 (close subclasses). Then, this noise could be due to the overlap effect instead of mislabeled issues. On the other hand, the column called "Noise Label" refers to noise between subclasses with a separation index larger or equal to 0.2 (well-separated subclasses). Therefore, noisy instances are likely associated with mislabeled issues. For the artificial training set, all noisy instances are associated with the overlap effect. Under the noise location concept, the diagnostic tool proposes three new metrics, which describes noise proportions for training sets. The first metric is the Noise Ratio (NR), which is the quantitative relation between valid and all noisy instances. For instance, the artificial training set has a NR = 17.14. That means there is one noise instance per 17.14 valid instances. The second metric is the Noise Overlap Ratio (NOR), that considers noisy instances associated with close subclasses. Finally, the Noise Label Ratio (NLR) considers noise related to well-separated subclasses. In the example, all noisy instances are associated with close subclasses. Therefore, NR = NOR, and NLR does not apply. Figure \protect\ref{FIG:5} shows the Noise matrix. This is a square matrix, whose size is determined by the number of subclasses. Each column of the matrix represents the instances in an estimated subclass based on the 3-Nearest Neighbors Rule while each row represents the instances in an actual subclass (original membership). Thus, the diagonal of the matrix contains the number of valid instances per subclass (instances that kept the same subclass label after the noise analysis). For instance, reading the first row: 176 instances from \textbf{N-01} did not change their membership, one instance moved to \textbf{N-03} (it is not considered noise because \textbf{N-01} and \textbf{N-03} belong to the same Negative class), 10 noisy instances from \textbf{P-01}, and 2 noisy instances from \textbf{P-02}.


\begin{table}[t]
\small
\begin{minipage}[t]{.38\textwidth}
   \centering
   \caption{Output 4: Noisy and valid instances per class for the artificial training set.}
   \label{tbl4}
   \begin{tabular}[t]{lll}
     \toprule
     Class & Noise & Valid \\
     \midrule
     Negative & 15 & 389 \\
     Positive & 12 & 74\\
     \bottomrule
   \end{tabular}
   
\end{minipage}\hfill
\begin{minipage}[t]{.58\textwidth}
   \centering
   \caption{Output 4: The number of noisy instances per subclass including the noise location for the artificial training set.
}\label{tbl5}
   \begin{tabular}[t]{llllll}
     \toprule
     & Instances & Valid & Noise & Noise & Noise \\
    Subclass & &  &  & overlap & label \\
     \midrule
      \textbf{N-01}& 189 & 177 & 12 & 12 & 0 \\
      \textbf{N-02}& 152 & 152 & 0 & 0 & 0 \\
      \textbf{N-03}& 63 & 60 & 2 & 3 & 0 \\
      \textbf{P-01}& 36 & 26 & 10 & 10 & 0 \\
      \textbf{P-02}& 50 & 48 & 2 & 2 & 0 \\
     \bottomrule
   \end{tabular}
\end{minipage}
\vspace*{1mm}
\end{table}

\item \textbf{Output 5: Dispersion test}. This component has two outputs. The first one presents the dispersion per subclass by estimating the average and standard deviation of the distances to the subclass median. For instance, Table  \protect\ref{tbl6} shows such dispersion per class. Specifically, \textbf{P-01} and \textbf{N-01} are the subclasses with larger dispersion values. Besides the dispersion table, this first output includes dispersion plots per class. Figure \protect\ref{FIG:6} presents such dispersions for the artificial training set, where visual sparseness not only coincides with the values from Table \protect\ref{tbl6}, but also with that observed in Figure \protect\ref{FIG:3}. In case of detecting two or more subclasses within a class (refer subclass detection component), a second output shows the results of the Anderson's multivariate test for homogeneity of variance. The goal is to test the null hypothesis that assumes variance is equal across subclasses. A \textit{p-value} less than 0.05 indicates a violation of the assumption. Therefore, same resampling or synthetic data treatments strategy cannot necessarily be generalized to all subclasses. For example, Table \protect\ref{tbl7} makes evident the latter statement for the artificial dataset.

\end{enumerate} 
\vspace*{2mm}



\begin{table}
\small
\begin{minipage}[t]{.3\textwidth}
   \centering
   \caption{Output 5: Subclass dispersion for the artificial training set.}
   \label{tbl6}
   \begin{tabular}[t]{lll}
    \toprule
    & Mean & Std  \\
    \textbf{N-01} & 0.96 & 0.46 \\
    \textbf{N-02} & 0.63 & 0.32 \\
    \textbf{N-03} & 0.65 & 0.33 \\
    \textbf{P-01} & 1.05 & 0.85 \\
    \textbf{P-02} & 0.57 & 0.41 \\
    \bottomrule
   \end{tabular}
   
\end{minipage}\hfill
\begin{minipage}[t]{.65\textwidth}
   \centering
   \caption{Output 5: Anderson's multivariate test for homogeneity of variance for the artificial training set. Df (Degree of freedom), Sum Sq (Sum of squares), Mean Sq (Mean squared), F (The F-test), N.Perm (Number of permutations), and \textit{p-value} (Probability value).}
   \label{tbl7}
   \begin{tabular}[t]{lcccccc}
     \toprule
     & Df & Sum.Sq & Mean.Sq & F & N.Perm & p-value \\
     \midrule
           \textbf{Negative}   & & & & & &\\
      Subclass           & 1  & 5  & 49  & 12.3 & 999 & \textbf{0.002}\\
      Residuals          & 84 & 34 & 0.4 &      &     & \\
      \textbf{Positive}   & & & & & &\\
      Subclass           & 2  &11  & 5.3  & 34 & 999 & \textbf{0.001}\\
      Residuals          & 401 &63 & 0.2 & & & \\
     \bottomrule
   \end{tabular}
\end{minipage}
\vspace*{3mm}
\end{table}



\begin{figure}
	\centering
		\includegraphics[scale=.33]{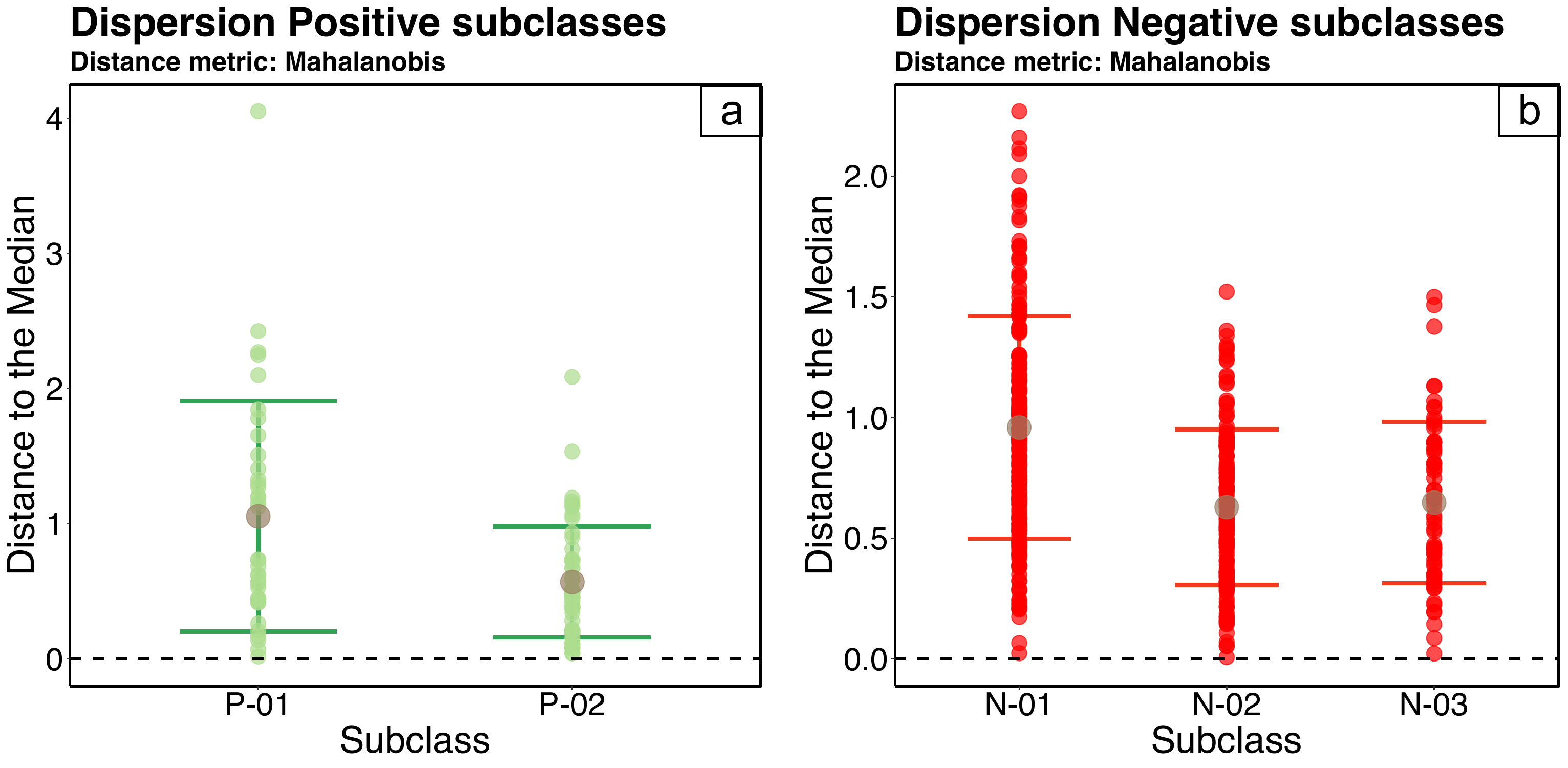}
	\caption{Output 5: Subclass dispersion for artificial training set. a) Dispersion for Negative subclasses.  b) Dispersion for Positive subclasses. 
The dotted lines represent the median as a reference point. For each subclass plotted on the X axis, the distance of each observation to the median, is plotted in the Y axis. Dark dots represent the mean of the distances to the median and the error bars the standard deviation.}
	\label{FIG:6}
\end{figure}


\newpage

\section{Experimental methodology}

Previous studies suggested that the class imbalance is not an issue by itself, but performance degradation of classifiers is also related to other problems such as overlapping, noisy labels, and small-disjuncts \cite{10.1007/978-3-540-76725-1_42,Prati2004b,Japkowicz2002,Jo2004,Prati2004a,Kubat1997,Batista2005}.  Although these papers indicate a relationship between such problems, the true correlation and causality between them is not yet well-established. Furthermore, common limitations of previous studies are based on lack of statistical validation, few datasets and classifiers, and the frequent use of simple or artificially generated datasets. Therefore, their conclusions may hold only for those limited scenarios.
The experimental framework implemented in the present research is composed of three stages:

\vspace*{2mm}
\begin{enumerate}
\itemsep=5pt
\item \textbf{DDP}: The first stage seeks to build "Dataset Degradation Profiles" (DDP) for datasets based on the diagnostic algorithm output. This makes possible to know deeper the relationship between degradation problems and their critical levels.

\item \textbf{Classifiers performance}: The second stage builds and evaluates several classification algorithms in order to assess the level of agreement between performance metrics, and tests classification algorithms robustness in presence of degradation problems.

\item \textbf{Treatments comparison}:  Finally, the last stage compares several techniques for handling degradation problems by making use of the DDP results. This stage mainly seeks to make evident that the success of classification tasks depends on the combination of several factors related to dataset diagnosis, treatment, and classification algorithm.

\end{enumerate} 
\vspace*{2mm}

The Diagnostic model and treatments (techniques for handling degradation problems) were studied empirically by using 49 datasets from real-world domains. Table \protect\ref{tbl14}   describes main characteristics of that collection, which varies in their size (instances), number of features (Feat.), and class imbalance. First 35 datasets are from the UCI machine learning repository \cite{Dua:2019}, 7 datasets, identified with `$\ast$', from \verb|R| packages, 6 datasets from different repositories, and one new dataset, the "Simulated dataset", generated by the author. Those multiclass datasets marked with `$\dagger$' were binarized using the original dataset description. Finally, datasets identified with `$\bullet$' are subsets of the original datasets. Experiments from second and third stages are based on 14 classification algorithms, whose parameter values were set based on two criteria:  authors recommendations and model tuning using resampling. In the latter case, the \verb|caret| (Classification And Regression Training) package \cite{Kuhn2008,Kuhn2008a,Kuhn2017,Kuhn2017a} was used, which is available in \verb|R| open source software. Package \verb|caret| offers a framework for building machine learning models. For instance, it includes several functions for data splitting, pre-processing, feature selection, variable importance estimation, model tuning using resampling. This research used that latter component, which allowed to choose the optimal model across parameters for classification algorithms. Finally, the experimental results are obtained based on 10-fold cross-validation. In summary, first stage results come from the 49 datasets diagnostic analysis. Second stage experimental results stem from every combination of the 49 datasets and 14 classifiers (674 trained models); in short, all classifiers in this stage were trained by using original raw datasets without the intervention of any remediation technique for handling degradation problems. Finally, last stage experimental results derive from every combination of the 49 datasets, 14 classifiers, and 9 treatments (5995 trained models). Table \protect\ref{tbl12} shows a brief description of the classifiers implemented, which lists functions, packages and tuning parameters for 14 classification algorithms. The list of treatments for degradation problems and set parameters are shown in Table \protect\ref{tbl13}.

\section{Experimental Results}

As previously mentioned, different studies suggest class imbalance is not completely responsible for classifiers degradation and argue that the problem of small-disjuncts is even more severe \cite{Prati2004b,Holte1989,Japkowicz2001,Nickerson2001,Jo2004,article}. However, the small-disjunct concept itself can be interpreted as the imbalance problem between-class and within-class. Furthermore, one of the reasons why small-disjuncts show higher misclassification than large disjuncts is due to class imbalance \cite{Weiss2003a,Japkowicz2003}. On the other hand, other studies state that misclassification is not solely caused by class imbalance, but also associated with the degree of overlapping among classes \cite{10.1007/978-3-540-76725-1_42,Prati2004b,Prati2004a}. Keeping the above in mind, the diagnostic tool integrates those elements by building an output called "IR and Overlap matrix", which creates a "diagnosis" or profile of the target dataset.

\subsection{Dataset profiling}

\begin{figure}[t]
	\centering
	    \includegraphics[scale=.5]{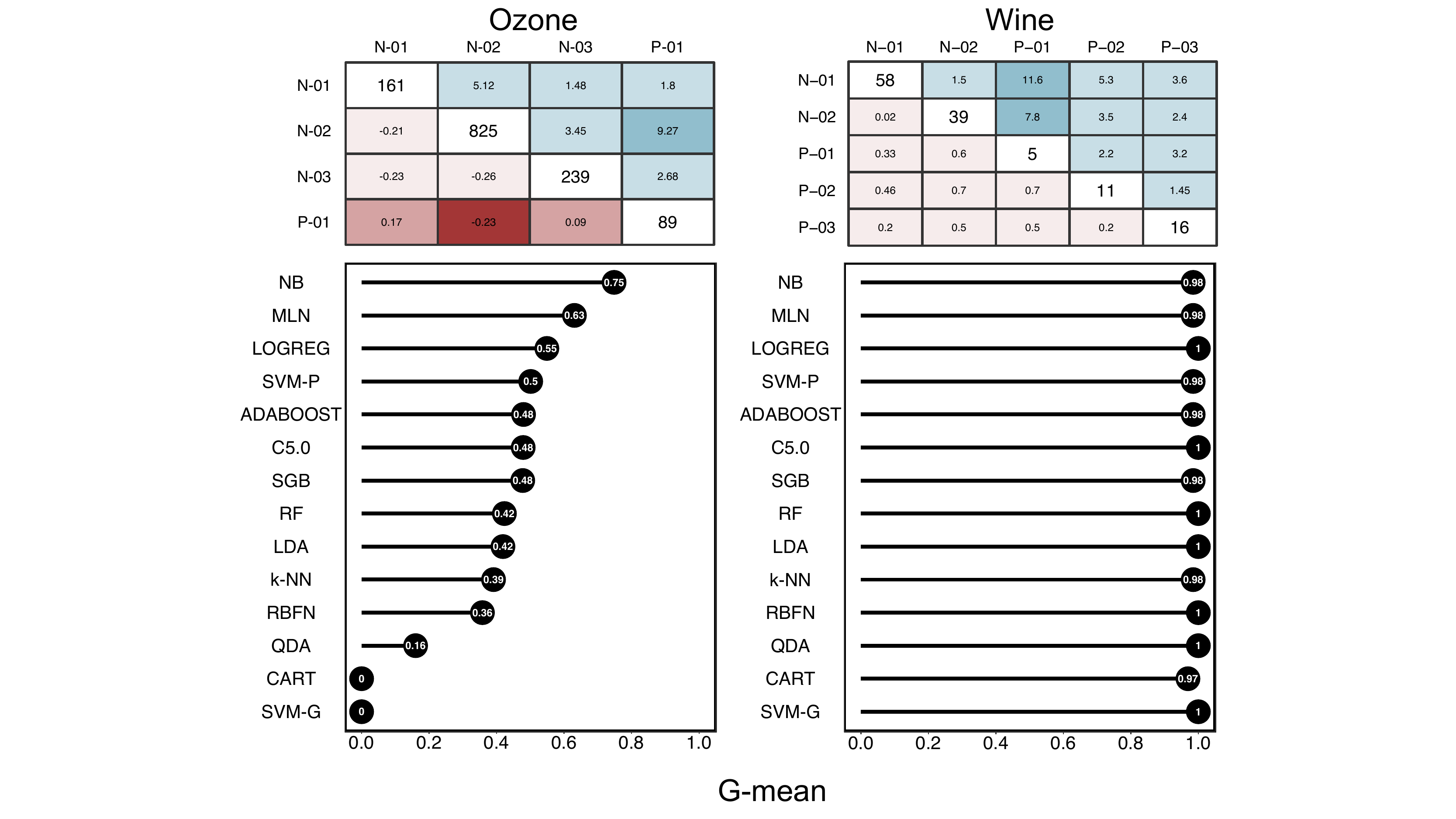}
	\caption{IRO matrix comparison and performance for multiple classifiers. The diagonal contains the number of instances per subclass (white), which allows for identifying small-disjuncts in both classes. The upper triangular part (blue) shows the IR between subclasses. Finally, the lower triangular part (red) of the matrix shows the separation index (degree of overlapping) between subclasses. The intensity of the colors identifies how critical the problem based on each metric is. The segment chart below of each IRO matrix shows the performance of 14 classifiers for each dataset by using the \textit{G-mean} performance metric.}
	\label{FIG:7}
\end{figure}

In the following, a comparative example is presented using the Ozone and the Wine datasets (Refer Table \protect\ref{tbl14}) to illustrate the benefits derived from the diagnostic model. Initially, the Ozone dataset looks more complex since it has higher dimensionality (number of features) and class imbalance (IR). On the other hand, the amount of data may be the major issue for the Wine dataset. Figure \protect\ref{FIG:7} compares the IR and Overlap matrix for those datasets. The diagonal of each matrix contains the number of instances per subclass (white), which allows for identifying small-disjuncts in both classes. The upper triangular part (blue) shows the IR between subclasses. Finally, the lower triangular part (red) of the matrix shows the separation index (degree of overlapping) between subclasses. The intensity of the colors identifies how critical the problem based on each metric is. It is important to mention that overlap between subclasses from the same class is not considered a problem.  Overall, Wine has more subclasses (disjuncts), less amount of data, higher IR between subclasses, and lack of overlap. In contrast, the most relevant problem of the Ozone dataset is related to several regions with high overlap and some subclasses with few instances. Moreover, the segment chart below of each IRO matrix shows the performance of 14 classifiers for each dataset by using the \textit{G-mean} performance metric. There is an evident low performance for most of the classifiers in the Ozone dataset case, suggesting a need to implement remediation treatments. Furthermore, even though the Wine dataset has more small-disjuncts and higher IR between subclasses, this dataset does not require any remediation treatment. Finally, the Wine dataset may be considered as not a challenging case for classification task given lack of overlapping between subclasses from different classes.

In general, the aim of the DDP (Dataset Degradation Profile) is to provide useful information for selecting or designing a tailored treatment. Dataset profiles are categorized according to levels of criticality as follows:

\vspace*{2mm}
\begin{itemize}
\itemsep=5pt
\item If IR is equal to or less than 10, then level is "Low", otherwise level is "High".
\item If the number of Disjuncts (subclasses) is equal to or less than 10, then level is "Low", otherwise level is "High".
\item If NOR (Noise Overlap Ratio) is equal or less than 0.1, then level is "Low", otherwise level is "High".
\end{itemize}
\vspace*{2mm}


\begin{figure}[t]
\begin{minipage}{.45\textwidth}
   \centering
   	\includegraphics[scale=.18]{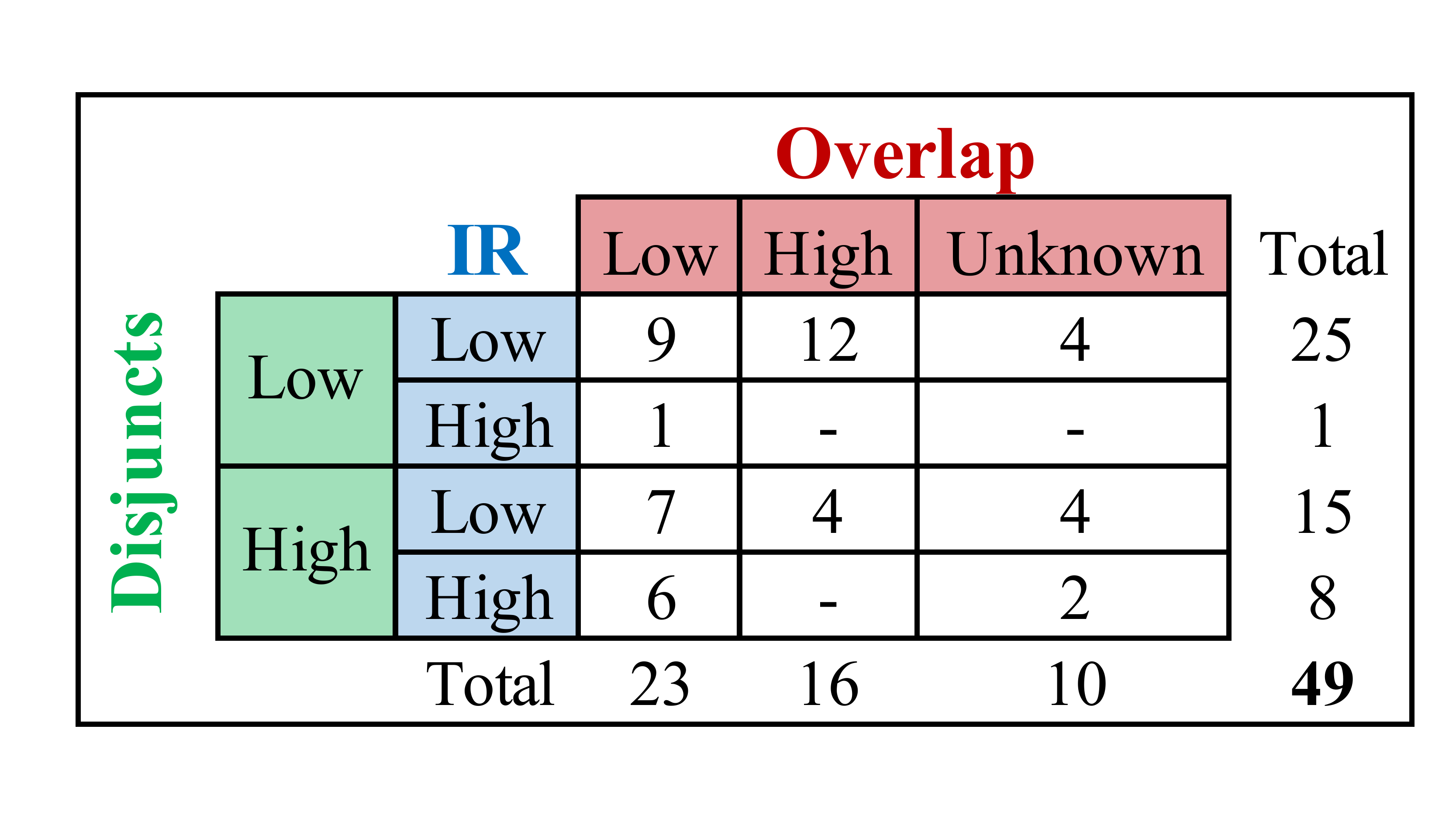}
	\caption{Contingency table of dataset profiles. Cross levels show the number of datasets for each criticality combination.}
	\label{FIG:8}
\end{minipage} \hfill
\begin{minipage}{.50\textwidth}
   \centering
   \includegraphics[scale=.33]{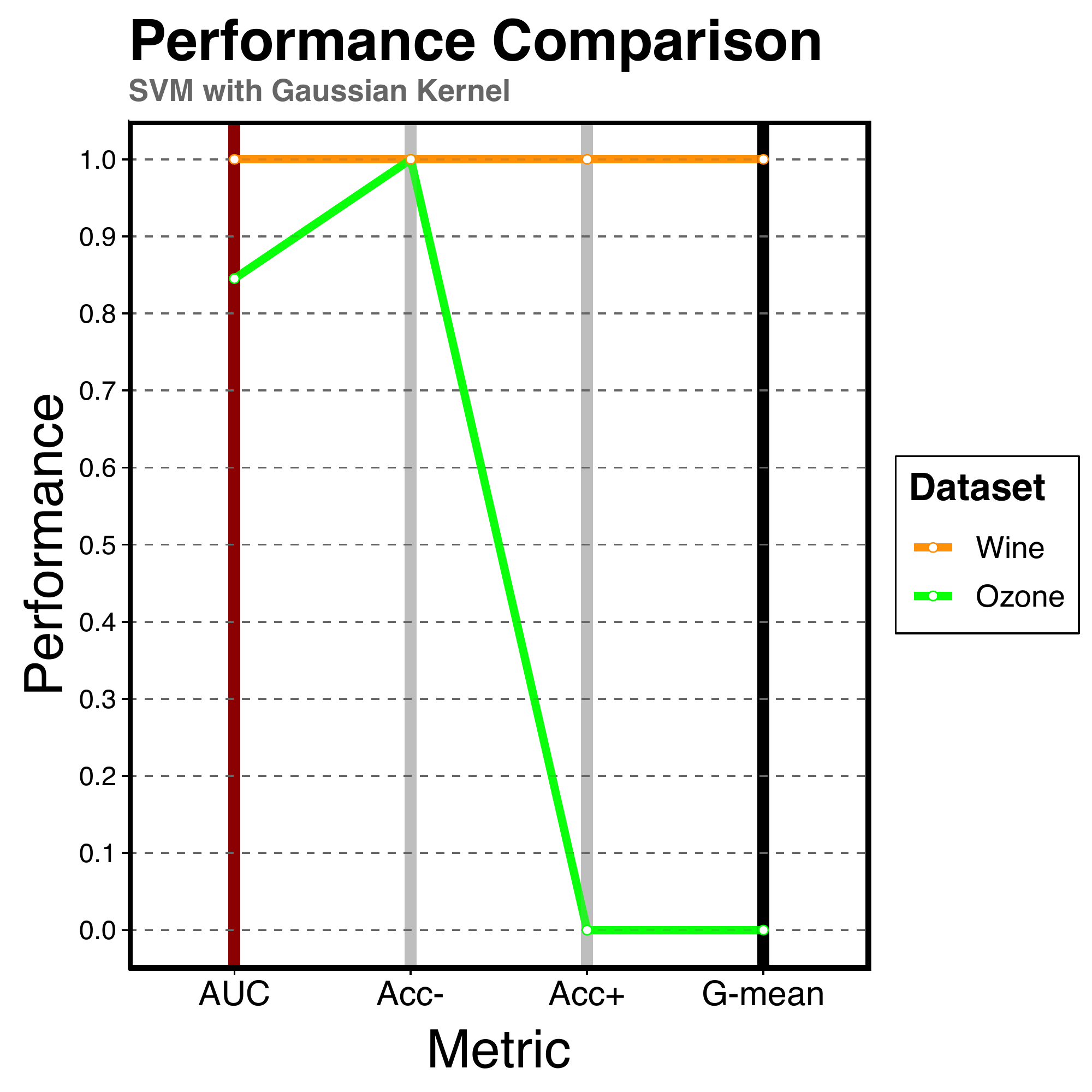}
	\caption{Performance metrics comparison by using SVM-G. The \textit{AUC} and the \textit{G-mean} are consistent with ${Acc}^-$ and ${Acc}^+$ for the Wine dataset. However, the \textit{AUC} is an optimistic metric for the Ozone dataset (85\%) despite its low ${Acc}^+$ (0\%). On the other hand, the \textit{G-mean} is more fair metric based on the levels of errors.}
	\label{FIG:9}
\end{minipage}
\end{figure}

Figure \protect\ref{FIG:8} shows a contingency table that summarize the results of the DDP process for the 49 datasets. The table describes the relationship between three degradation problems (IR, Disjuncts, and Overlap) in terms of frequencies of their levels of criticality (Low and High). The level "Unknown" from the Overlap problem refers to datasets for which it was not possible to find an optimal projection for estimating the separation index.

\newpage


\subsection{Performance Metrics Selection}

Performance metrics are used to evaluate how well classification algorithms conduct the discrimination task between classes. Several studies point out limitations of the global accuracy as a metric of performance in class imbalance situation \cite{Weiss2004,Maloof2003,Provost1997a,Provost1998,He2009,Weiss2001}. In a nutshell, the global accuracy metric is heavily biased to favor the Negative class and may hide poor performance for the Positive class. Furthermore, Positive class is usually the class of interest in real world applications. Typically, the aim is to learn in detail the Positive class distribution in order to recognize its members as accurately as possible. For example, fraud detection is a well-known case in point, where the number of fraud transactions is much smaller than the legitimate ones and the goal is to detect such a transactions immediately in order to avoid financial losses. In conclusion, the main goal of classifiers is not only improving the accuracy for the Positive class, but also avoiding decreasing the accuracy for the Negative class during the process. Continuing with the same fraud example, financial institutions want to increase the detection of the number of "True Positive" (\textit{TP}) cases  (real fraud transactions) and avoid increasing the number of "False Positive" (\textit{FP}) cases (legitime transactions classified as fraud) because the latter means higher administrative costs related to responding to the false fraud alerts. Practitioners have opted to use the Area under the Curve \textit{AUC} of the Receiver Operating Characteristic \textit{ROC} \cite{Swets1988} as the metric of performance for supervised learning, since it is considered more appropriate for imbalance learning \cite{Bradley1997,Fawcett2004,Fawcett2006}. However, recent studies have shown that  the \textit{AUC} may provide an optimistic performance evaluation \cite{Davis2006, Saito2015} and draw wrong conclusions for imbalance datasets. Thereby, the first crucial contribution of this study in this context is to clarify how convenient it is to use the \textit{AUC} as performance metric for datasets in the presence of degradation problems. If the \textit{AUC} metric is shown to be optimistic and misleading, it will be not only inconvenient for classifiers performance evaluation, but also for testing the effectiveness of the remediation treatments for degradation problems such as SMOTE, OSS, ENN, and so on. For instance, Figure \protect\ref{FIG:9} compares four performance metrics (\textit{AUC}, ${Acc}^-$, ${Acc}^+$, and \textit{G-mean}) associated with a Support Vector Machine with Gaussian Kernel (SVM-G) implementation for the Ozone and the Wine datasets.

\vspace*{3mm}

As previously mentioned, the Wine dataset is not a challenging case for classification; Thereby, these metrics agree in terms of the SVM-G. Nevertheless, the Ozone dataset shows discrepancy between some metrics. Even though the \textit{AUC} is above 80\% and the accuracy of the Negative class (${Acc}^-$) is 100\%, the accuracy of the Positive class (${Acc}^+$) is 0\%. This could imply that the \textit{AUC} is an inaccurate metric in relation to certain cases. For example, the low ${Acc}^+$ may be associated with high level of global IR (13.76) in the Ozone dataset. In summary, there is one instance in the Positive class per 13.76 instances in the Negative class. Moreover, the separation index shows overlap problems between subclasses (IRO matrix from Figure 6.2). On the other hand, the \textit{G-mean} is more conservative and intuitive metric because shows the balance between the accuracy from both classes. Since the ${Acc}^+$ for the Ozone dataset is zero, the \textit{G-mean} punishes the classifier performance because it is unable to classify one of the classes correctly. Technically, the \textit{G-mean} (also called \textit{G-measure}) is the geometric mean of the ${Acc}^+$ and the  ${Acc}^-$, which returns a value between 0 and 1. Technical details of the evaluation metrics implemented in the present research such as \textit{AUC}, (\textit{F1-score}), \textit{G-mean}, \textit{sensitivity}, \textit{specificity}, and \textit{precision} can be found in Appendix B. This example makes evident limitations of the \textit{AUC} as performance metric for complex datasets. Since the \textit{ROC} graphs are drawn using True Positive Rates (\textit{TPR}) and False Positive Rates (\textit{FPR}), the relative imbalance between classes directly affects the \textit{FPR}, which tends to zero and generates misleading estimates of the \textit{AUC} measure.

\begin{figure}[t]
\begin{minipage}[t]{.45\textwidth}
   \centering
   	\includegraphics[scale=.5]{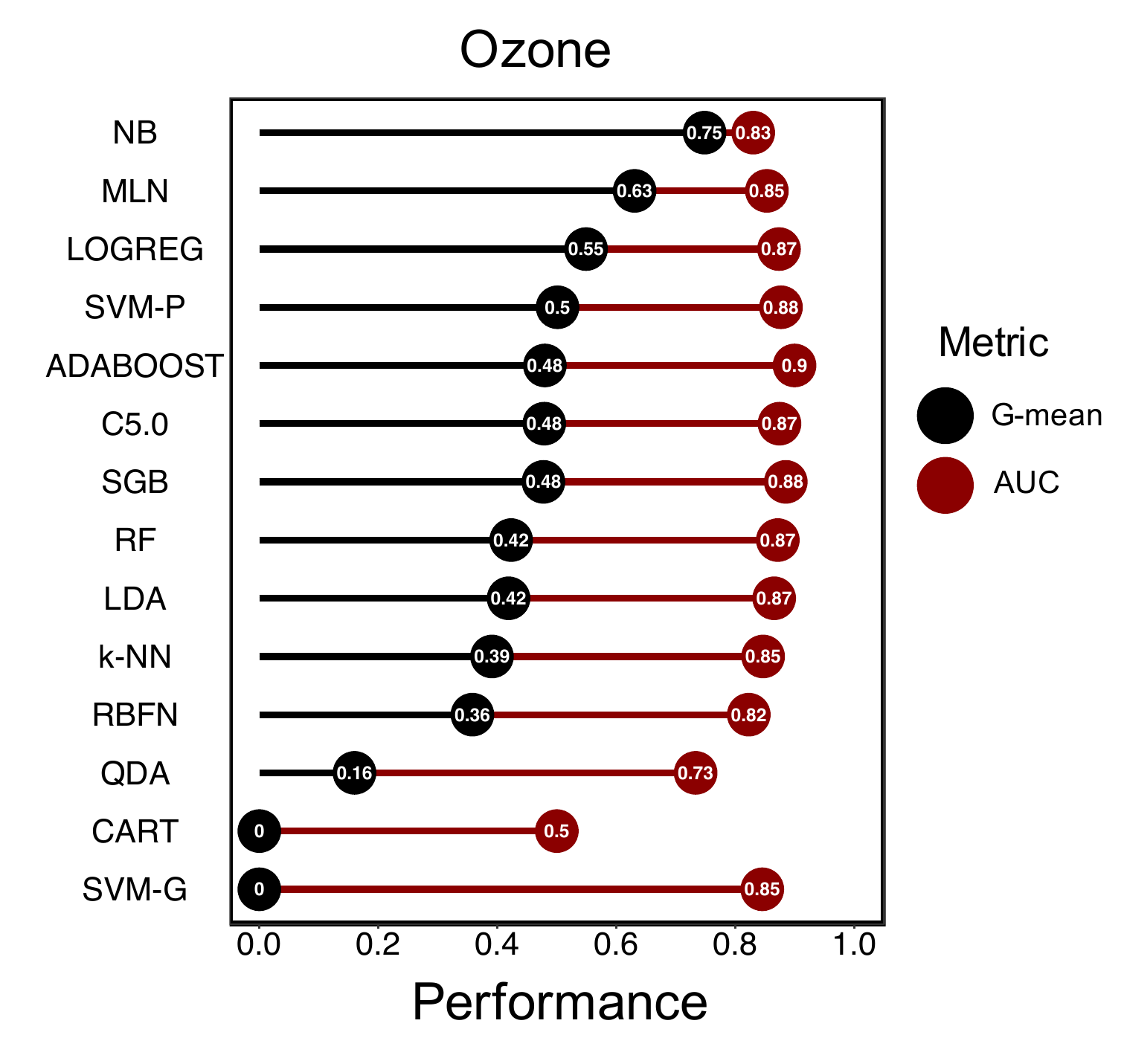}
	\caption{Performance metric comparison for the Ozone dataset across 14 classification algorithms. The \textit{AUC} for all classifiers get higher performance values (more optimistic) than the \textit{G-mean}.}
	\label{FIG:10}
\end{minipage}\hfill
\begin{minipage}[t]{.5\textwidth}
   \centering
   \includegraphics[scale=.32]{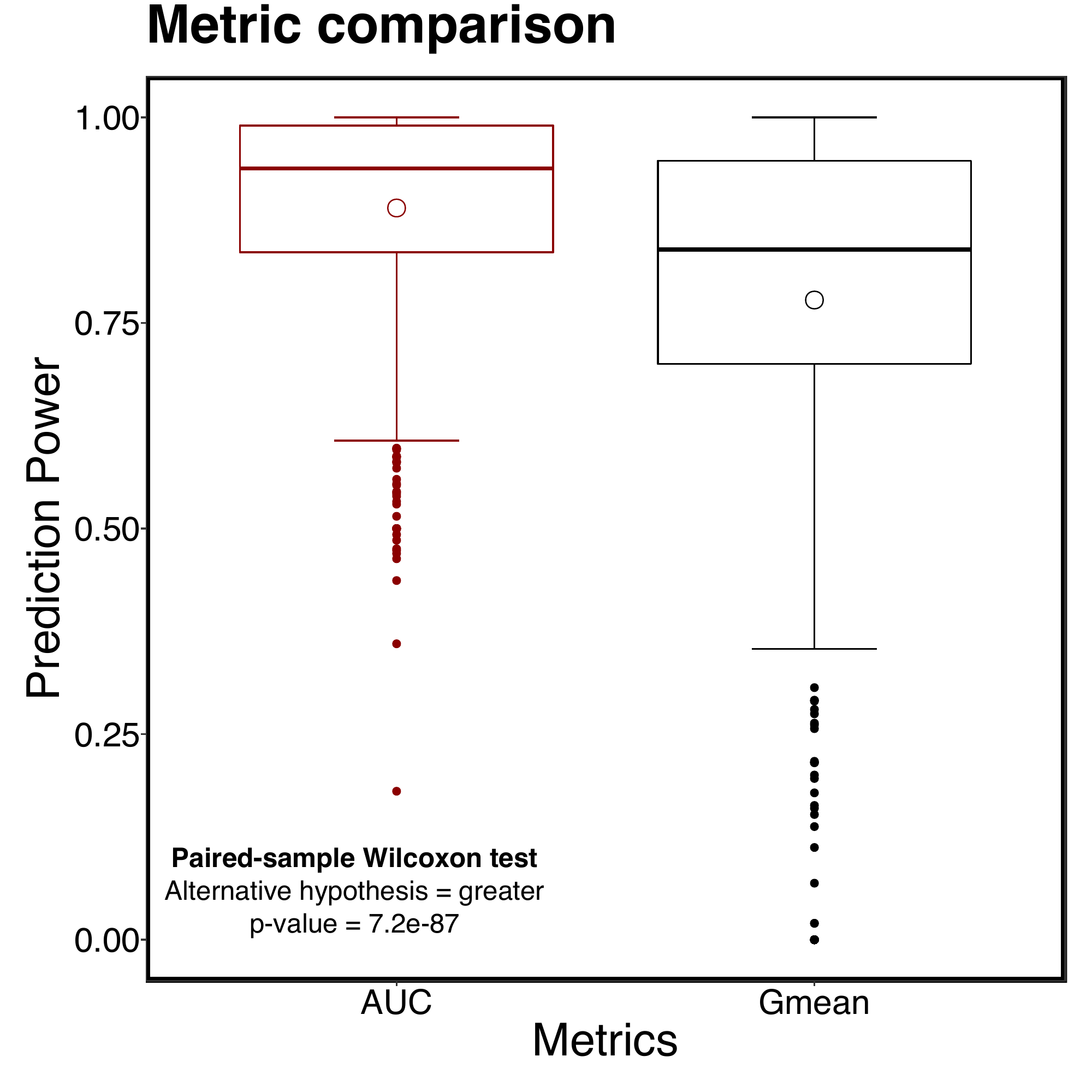}
	\caption{Boxplots for \textit{AUC} and \textit{G-mean} metrics. It compares the performance distributions of the 49 datasets across the 14 classification algorithms. The \textit{AUC} shows more optimistic prediction power than the \textit{G-mean}. Moreover, the paired samples Wilcoxon test rejects the hypothesis that states the \textit{AUC} has values of prediction power equal to or lower than the \textit{G-mean}.}
	\label{FIG:11}
\end{minipage}
\end{figure}

In general, most classifiers are not designed with the assumption of dealing with class imbalance \cite{Chawla2004}, Figure \protect\ref{FIG:9} is a clear example of this by using ${Acc}^+$ and the \textit{G-mean} metrics, which are performance measures more conservative than the \textit{AUC}. The SVM-G showed a low performance in term of ${Acc}^+$ for the Ozone dataset, which is not surprising due to its high imbalance ratio. Figure \protect\ref{FIG:10} clearly shows the disagreement between \textit{AUC} and \textit{G-mean} metrics for the Ozone dataset across 14 classification algorithms, where the \textit{AUC} for all cases always overpass the \textit{G-mean}. Another important finding related to the Wine dataset is that for well-separated subclasses the performance metric did not decrease even with high levels of imbalance ratio between subclasses. Thereby, the present experimental stage has two main purposes: the first one seeks to generalize the improper use of the \textit{AUC} as performance metric for datasets under degradation problems. The second one refers to establishing solid causality with statistical evidence between degradation problems and classifiers' performances by using real-world datasets.\

Figure \protect\ref{FIG:11} compares the distribution of the \textit{AUC} and \textit{G-mean} metrics for the 49 datasets across the 14 classification algorithms (674 trained models). The \textit{AUC} boxplot shows a lower variability and its values are closer to the high levels of prediction power. Moreover, \textit{AUC} outliers below the lower quartile show better performance than the outliers from the \textit{G-mean}. Finally, the measures of central tendency such as the median and the mean have larger values in the \textit{AUC} than the \textit{G-mean} distribution. Furthermore, this figure shows the result of the paired samples Wilcoxon test \cite{Mann1947,Hollander1999,Conover1999}, which is a non-parametric version of the paired t-test. The null hypothesis states that the \textit{AUC} has values of prediction power equal to or lower than the \textit{G-mean} values. In contrast, the alternative hypothesis claims that the \textit{AUC} has values of prediction power higher than the \textit{G-mean} values. As the \textit{p-value} turns out to be 5.06e-88, and is less than the 0.05 significance level, it is possible to reject the null hypothesis. Therefore, for the experimental design defined in the present research, the Wilcoxon for paired samples test confirms the risk and inconvenience of using the \textit{AUC}. A fair question could be raised as to the \textit{AUC} limitations being the same through levels of criticality of the degradation problems.


\newpage

\begin{figure}
	\centering
		\includegraphics[scale=.43]{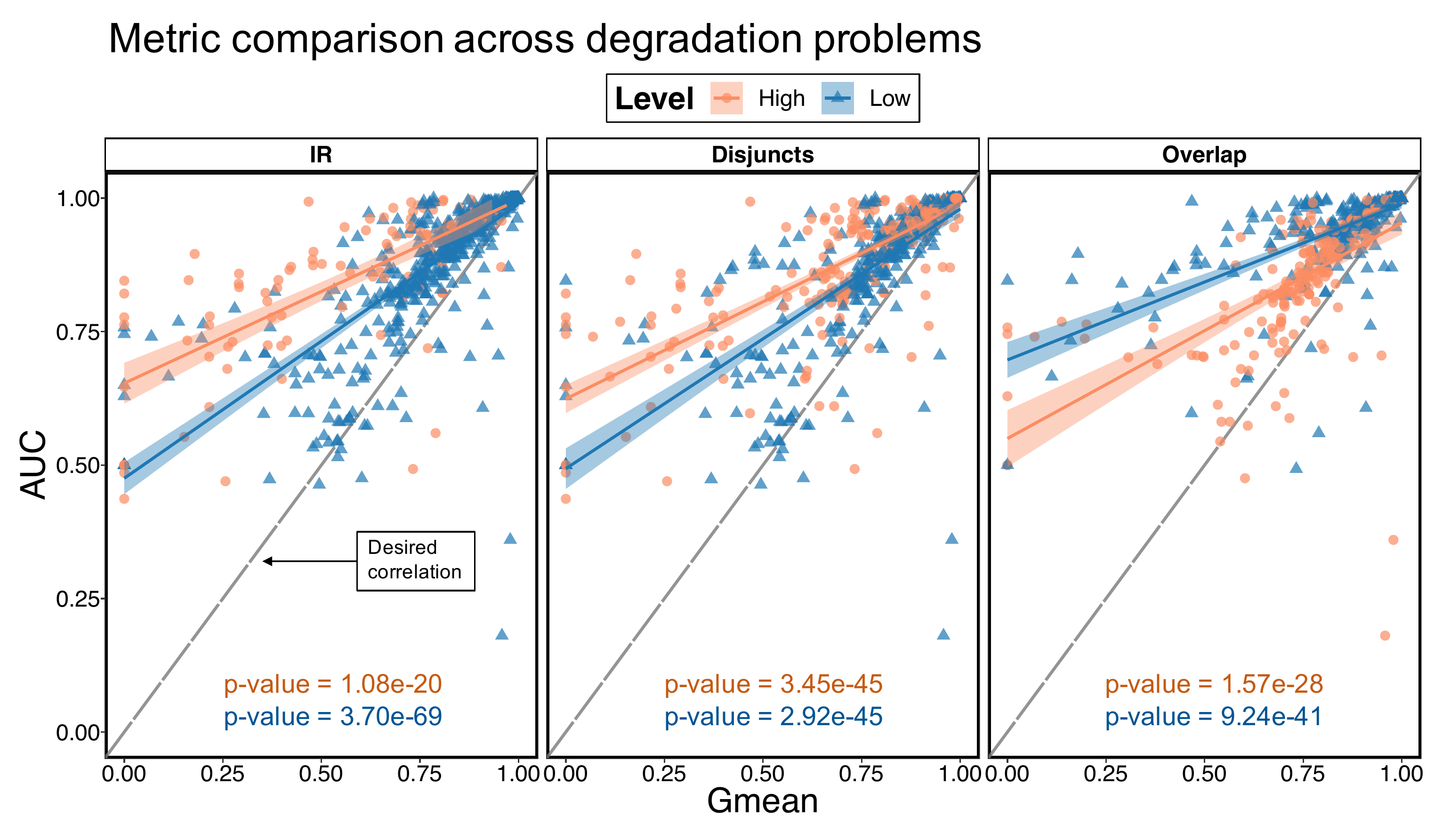}
	\caption{\textit{AUC} and \textit{G-mean} comparison across criticality levels of degradation problems. It compares the scatter between \textit{AUC} and \textit{G-mean} by levels of criticality of degradation problems for the 49 datasets across the 14 classification algorithms. In a nutshell, if the dots do not lie close to the 45-degree line, discrepancy between metrics exists. Moreover, the paired samples Wilcoxon test rejects the hypothesis that states the \textit{AUC} has values of prediction power equal to or lower than the \textit{G-mean} values.}
	\label{FIG:12}
\end{figure}

Figure \protect\ref{FIG:12} answers the above question by comparing the scatter between \textit{AUC} and \textit{G-mean} for levels of criticality of each degradation problem. The ideal scenario would be a high linear positive correlation between values for both performance metrics, even for the low and high values. In a nutshell, if the dots do not lie close to the 45-degree line, discrepancy between metrics exists. It is evident that the \textit{AUC} limitations persist through the different levels of criticality; however, for high levels of IR, the \textit{AUC} is more optimistic due to the low values of the \textit{FPR}. For the next degradation problem, the dispersion of the disjunct levels does not show significant difference. Finally, the overlap problem shows an interesting behavior through its levels. High levels of overlap make harder to learn the distribution for both classes. Then, the \textit{FPR} is increased and the \textit{TPR} is reduced, which decreases the \textit{AUC} values. On the other hand, low levels of overlap avoid decreasing the "True Negative" value and follow a similar dispersion due to the high level of IR.
Similarly, the paired samples Wilcoxon test was conducted for each level of criticality. Since all \textit{p-values} turns out to be less than the 0.05 significance level, it is possible to reject the null hypothesis. Therefore, for the experimental design defined in the present research, the Wilcoxon for paired samples test confirms the risk and inconvenience of using the \textit{AUC} even for different levels of criticality.


\subsection{Classifiers robustness in the presence of degradation problems}

Once \textit{AUC} limitations in the presence of degradation problems are evidenced, the next aim of this research is to test classification algorithms$'$ robustness in the presence of degradation problems. The following results correspond to the same experimental design defined in the previous section. Basically, this section seeks to validate three recurrent statements in the literature by incorporating a considerable real-world dataset, several classification algorithms, unbiased performance metrics, and statistical validation.\
The three statement to validate are:

\vspace*{2mm}
\begin{itemize}
\itemsep=5pt
\item[\textbf{\textit{S1:}}] The more class imbalance, the bigger the classifier performance degradation.
\item[\textbf{\textit{S2:}}] The greater the number of disjuncts, the bigger the classifier performance degradation.
\item[\textbf{\textit{S3:}}] The larger the overlap regions, the bigger the classifier performance degradation.
\end{itemize} 
\vspace*{2mm}

\newpage

Table \protect\ref{tbl8} illustrates the results of an alternative version of the Wilcoxon signed-rank test, which is used to compare two related samples. In this example, results are derived with the 49 datasets and SVM-G classifier by making use of three performance metrics (\textit{G-mean}, \textit{F1-score}, and \textit{AUC}). Basically, two tests were implemented in order to validate if statistical differences between the performance metric distribution across the levels of criticality for each degradation problem exist. In other words, these tests assess the truthfulness of the three above statements. Test A compares if the performance metric distributions between levels are equal, and Test B compares if the performance metric distribution of the datasets with high levels of criticality have better prediction power than the performance metric distribution of the datasets with low levels of criticality. Both tests have a 0.05 significance level. The \textit{p-values} shown in the table are used to reject or not reject the null hypothesis. However, an easier notation combines results of both statistical tests. Symbol "$\checkmark$" means existing statistical evidence to accept the \textbf{statement}. On the other hand, symbol 	"$\times$" implies that there is no statistical evidence to accept the \textbf{statement}. Therefore, it is not possible to reject that the performance metric distributions across the levels are different.


\begin{table}
\small
\caption{\textit{p-values} of the Wilcoxon signed-rank tests for degradation problems by using a SVM-G classifier.
Test A compares if the performance metric distributions between levels are equal and Test B compares if the performance metric distribution of the datasets with high levels of criticality have better prediction power than the performance metric distribution of the datasets with low levels of the criticality.}\label{tbl8}
 \centering
 
 \begin{tabular}[t]{lcccccc}
    \toprule
    &\multicolumn{6}{c}{\textbf{Statements for degradation problems}}\\
Metric      & \multicolumn{2}{c}{\textbf{\textit{S1}} (IR)}  &  \multicolumn{2}{c}{\textbf{\textit{S2}} (Disjuncts)}   & \multicolumn{2}{c}{\textbf{\textit{S3}} (Overlap)} \\
     \midrule
\textbf{G-mean}       & &       & &     & &        \\
\midrule
Test A       & 0.00812  & \multirow{2}{*}{$\checkmark$}   & 0.86476   & \multirow{2}{*}{$\times$}    & 0.07202  & \multirow{2}{*}{$\checkmark$}        \\
Test B       & 0.00406  &                        & 0.57548   &                         & 0.03601  &                            \\
\midrule
\textbf{F1-score}       & &       & &     & &        \\
\midrule
Test A       & 0.03484  & \multirow{2}{*}{$\checkmark$}   & 0.82594   & \multirow{2}{*}{$\times$}  & 0.06566    & \multirow{2}{*}{$\checkmark$}      \\
Test B       & 0.01742  &                        & 0.41297   &                       & 0.03283    &                           \\
\midrule
\textbf{AUC}       & &       & &     & &        \\
\midrule
Test A       & 0.15913  & \multirow{2}{*}{$\times$}    & 0.984    & \multirow{2}{*}{$\times$}  & 0.01767    & \multirow{2}{*}{$\checkmark$}       \\
Test B       & 0.07956  &                         & 0.51598  &                       & 0.00883    &                             \\
    \bottomrule
    \end{tabular}
\end{table}


Analyzing the outputs from Table \protect\ref{tbl8}, it is possible to state the following conclusions for the classifier SVM-G under the experimental framework defined in this research:

\vspace*{2mm}
\begin{enumerate}
\itemsep=5pt
\item The more the class imbalance, the bigger the classifier performance degradation for metrics \textit{G-mean} and \textit{F1-score}. On the other hand, \textit{AUC} metric is not affected by the imbalance problem since it is an optimistic and misleading performance metric.
\item It is not true that the greater the number of disjuncts is, the bigger the classifier performance degradation occurs; however, disjuncts detection procedures allow to estimate the imbalance ratio within classes \cite{article,Weiss2003a}.
\item It is true that the larger the overlap regions, the bigger classifier performance degradation.
\end{enumerate} 
\vspace*{2mm}

Table \protect\ref{tbl9} shows the same statistical analysis for the complete set of classifiers by using the \textit{G-mean} as performance metric, which is a more conservative and intuitive metric than the \textit{AUC} because it shows the balance between the accuracy from both classes. The column "Tests" validates the statements about level of criticality (Test A and Test B) and column "\% Diff" shows the difference in percentage between the means of the levels of criticality. Some important findings are the followings:

\vspace*{2mm}
\begin{itemize} 
\itemsep=5pt
\item Ten out of fourteen classifiers present \textit{G-mean} degradation in the presence of high levels of IR. 
\item Thirteen of fourteen classifiers show \textit{G-mean} degradation in the presence of high levels of overlap.
\item The \textit{G-mean} for all classifiers is not affected by high levels of disjuncts.
\item QDA classifier does not show statistical difference for the \textit{G-mean} distribution between high and low levels of IR, disjuncts and overlap.
\end{itemize}
\vspace*{2mm}



\begin{table}
\small
\caption{Wilcoxon test for degradation problems across classifiers by using the \textit{G-mean} performance metric. The column "Tests" validates the statements about level of criticality and column "\% Diff." shows the difference in percentage between the means of criticality levels.}
\label{tbl9}
 \centering
 \begin{tabular}{lccccccc}

    \toprule
    &\multicolumn{6}{c}{\textbf{Statements for degradation problems}}\\
    & \multicolumn{2}{c}{\textbf{\textit{S1}} (IR)}  &  \multicolumn{2}{c}{\textbf{\textit{S2}} (Disjuncts)}   & \multicolumn{2}{c}{\textbf{\textit{S3}} (Overlap)} \\
     \midrule
\textit{Classifier}&\textbf{Tests}  & \textbf{\%Diff.}    & \textbf{Tests}     & \textbf{\%Diff.}    & \textbf{Tests}   &\textbf{\%Diff.} \\
NB        & $\times$       & 0.07      & $\times$    & 0.06    & $\checkmark$   & 0.05 \\
LOGREG    & $\checkmark$   & 0.15      & $\times$    & 0.03    & $\checkmark$   & 0.17 \\      
LDA       & $\times$       & 0.13      & $\times$    & 0.03    & $\checkmark$   & 0.17  \\
QDA       & $\times$       & 0.12      & $\times$    & -0.01   & $\times$       & 0.02  \\
C5.0      & $\checkmark$   & 0.17      & $\times$    & -0.05   & $\checkmark$   & 0.12  \\
CART      & $\checkmark$   & 0.21      & $\times$    & -0.08   & $\checkmark$   & 0.14  \\
k-NN      & $\checkmark$   & 0.22      & $\times$    & -0.03   & $\checkmark$   & 0.10  \\
MLN       & $\times$       & 0.16      & $\times$    & 0.00    & $\checkmark$   & 0.14  \\
SVM-G     & $\checkmark$   & 0.29      & $\times$    & -0.01   & $\checkmark$   & 0.09  \\
SVM-P     & $\checkmark$   & 0.27      & $\times$    & 0.05    & $\checkmark$   & 0.18  \\
RF        & $\checkmark$   & 0.23      & $\times$    & 0.01    & $\checkmark$   & 0.08  \\
SGB       & $\checkmark$   & 0.18      & $\times$    & 0.01    & $\checkmark$   & 0.09  \\
ADABOOTS  & $\checkmark$   & 0.23      & $\times$    & 0.01    & $\checkmark$   & 0.08  \\
RBFN      & $\checkmark$   & 0.24      & $\times$    & -0.02   & $\checkmark$   & 0.08  \\

    \bottomrule
    \end{tabular}
\end{table}


\subsection{Statistical identification of the best treatment}

To demonstrate that it is not possible to find one treatment that is always the best in remediation for all degradation problems and classifiers, a ranking score was computed over three performance metrics (\textit{AUC}, \textit{F1-score}, and \textit{G-mean}) to each treatment results across the 5995 trained models, which were built over valid combination of 49 datasets, 14 classifiers, and 9 treatments (techniques for handling degradation problems). Therefore, the ranking score for the best performing treatment is 9 and the worst performing treatment is 1. Then, the nonparametric Friedman test \cite{doi:10.1080/01621459.1937.10503522,Friedman1939,CIS-289548} and multiple comparison of treatments were applied to the ranking results. This test not only allows to detect differences between the treatments results across multiple experimental related samples, but also implements the post hoc Friedman tests (multiple comparisons between treatments) by using the criterion Fisher$'$s Least Significant Differences (LSD) \cite{Hollander1999,Conover1999,Fisher1935}, which groups the treatments according to similarities by using the Compact Letter Display method \cite{Piepho2004}. The null hypothesis for the Friedman test is that there are no differences between treatment results across multiple experimental attempts. If the null hypothesis is rejected (\textit{p-value} less that the 0.05 significance level), it can be concluded that at least two of the treatments are significantly different from each other. Then, a multiple comparisons analysis can be executed in order to know which of the treatments are significantly different from which other treatments.  In particular, the multiple comparisons results are presented by listing the treatments in order of decreasing average score and grouping the treatments that are not significantly different. Thus, the procedure creates groups (identified by letters) as a union of different treatments that can contain one or more members, and the members of these groups are the nine different treatment in this research (Raw, Random, SMOTE, B-SMOTE, DBSMOTE, ADASYN, ENN, NCL, and OSS). Treatments with the same letter(s) are not statistically different. On the contrary, treatments that are statistically different get different letters. Treatments can have more than one letter to reflect overlap between the groups of treatments. Table \protect\ref{tbl10} shows the Friedman test results for 49 datasets, 14 classifiers, and 9 treatments by using the \textit{G-mean} performance metric.  According to the \textit{p-value} column, all values turns out to be less than the 0.05 significance level. Therefore, it is possible to reject the null hypothesis and accept that there are statistical differences between the treatments across classifiers. For instance, the NB (Naive Bayes) classifier shows better performance by using SMOTE treatment (letter "\textbf{a}"). On the contrary, B-SMOTE, DBSMOTE, NCL, and OSS treatments grouped by the letter "\textbf{c}" are the less convenient alternatives for the NB classifier; even worse than the Raw data (original data). These results call into question those in previous research on degradation problems, which present successful techniques across few classifiers and datasets. More importantly, this analysis supports one of the hypotheses of the present research that states it is not possible to find one treatment that is the best in remediation for all degradation problems, datasets or classifiers. The selection of the "best treatment" or even the most convenient classification algorithm depends on the available information, knowledge associated with the target dataset, and strengths and weaknesses of classification algorithms. For this reason, the diagnostic test for degradation problems has a direct relevance to the treatment selection in order to obtain successful classification outcomes and to avoid degradation.\


\begin{table}[t]
\small
\caption{Friedman test results for all datasets by using the \textit{G-mean} metric. According to the \textit{p-value} column, 
all values turns out to be less than the 0.05 significance level. Then, there are statistical differences between treatments across classifiers.}
\label{tbl10}
 \centering
 \begin{tabular}{|r|ccccccccc|c|}
    \toprule
    \hline
&\multicolumn{9}{c|}{\textbf{Treatment}} &\\
\textbf{Classifier}& \rotatebox{90}{Raw}  & 
\rotatebox{90}{Random}    & \rotatebox{90}{SMOTE}    & \rotatebox{90}{B-SMOTE}    & \rotatebox{90}{DBSMOTE}   & \rotatebox{90}{ADASYN} & \rotatebox{90}{ENN} & \rotatebox{90}{NCL} & \rotatebox{90}{OSS} & \textbf{\textit{p-values}} \\
\cline{2-10}
NB        & bc  & ab  & a    & c   & c     & bc   & abc  & c   & c    & $2.0\mathrm{e}{-02}$ \\
LOGREG    & e   & a   & ab   & bc  & ab    & bc   & cd   & d   & de   & $2.2\mathrm{e}{-10}$ \\      
LDA       & e   & a   & bc   & bc  & cd    & ab   & e    & de  & e    & $1.0\mathrm{e}{-10}$ \\    
QDA       & d   & ab  & abc  & cd  & bcd   & a    & abcd & ab  & abcd & $2.2\mathrm{e}{-02}$ \\    
C5.0      & d   & ab  & ab   & a   & abcd  & a    & cd   & bcd & abc  & $4.6\mathrm{e}{-04}$ \\    
CART      & d   & a   & ab   & ab  & ab    & a    & cd   & ab  & bc   & $8.1\mathrm{e}{-05}$ \\    
k-NN      & c   & a   & ab   & bc  & ab    & a    & c    & ab  & ab   & $4.4\mathrm{e}{-06}$ \\    
MLN       & c   & ab  & ab   & ab  & bc    & a    & bc   & ab  & bc   & $2.1\mathrm{e}{-02}$ \\    
SVM-G     & e   & a   & a    & ab  & bc    & a    & d    & bc  & c    & $0.0\mathrm{e}{-00}$ \\    
SVM-P     & e   & ab  & a    & bc  & ab    & ab   & de   & bc  & cd   & $3.5\mathrm{e}{-09}$ \\    
RF        & d   & b   & a    & b   & b     & a    & cd   & bc  & bc   & $1.3\mathrm{e}{-08}$ \\    
SGB       & e   & a   & ab   & ab  & bc    & ab   & de   & bcd & cd   & $2.2\mathrm{e}{-07}$ \\    
ADABOOTS  & d   & bc  & ab   & abc & ab    & a    & cd   & bc  & abc  & $6.3\mathrm{e}{-06}$ \\    
RBFN      & e   & cd  & a    & cd  & bc    & ab   & de   & bc  & c    & $1.1\mathrm{e}{-07}$ \\    

\hline
    \bottomrule
    \end{tabular}
\end{table}

\newpage

As previously mentioned in the experimental methodology, one of the goals is to compares several treatments for degradation problems by making use of the DDP results over 49 datasets. The new hypothesis states that success of classification tasks depends on the combination of several factors such as DDP, treatment, and classification algorithm. Based on the information from Figure \protect\ref{FIG:8}, two dataset profiles are selected for validating the hypothesis:

\vspace*{2mm}
\begin{itemize} 
\itemsep=5pt
\item \textbf{Profile A}: datasets with high levels of class imbalance and levels of disjuncts. 8 datasets have this profile: Seismic, Wilt, Abalone, Blocks, nanoHUB, Fraud, Mammo, and SDSS.
\item \textbf{Profile B}: datasets with high levels of overlap and low levels of class imbalance. 16 datasets have this profile:  Blood, College, Glass, ILPD, Ionosphere, MDRR, Phoneme, Pima1, Pima2, Ringnorm, Satimage, Sonar, Spam, Sports, Vertebral, and Weather.
\end{itemize}
\vspace*{2mm}

Similarly to the results for the complete datasets, the Friedman test for these two profiles showed statistical differences between the treatments across the 14 classifiers by using the mean of  the \textit{G-mean} ranks. On the contrary, if this test uses the mean of the \textit{AUC} ranks; then, there is no statistical differences between the treatments across several classifiers for both profiles. This situation not only makes evident the drawback of the \textit{AUC} mentioned in the sections before, but also its limitations as performance metric for comparing remediation treatments for degradation problems. Table \protect\ref{tbl11} shows the heat maps of the mean rank for the \textit{G-mean} metric for both dataset profiles across treatments and classifiers. High mean rank values tend to increase the steepness of the color gradient to green. On the the other hand, low mean rank values tend to decrease the steepness of the color gradient to red. The purpose of these figures is not to identify the best classification algorithm, but the best treatment in terms of the DDP and classifier. It is evident Profiles A and B show different response patterns.


\clearpage

In general, training sets treated with remediation techniques showed better classification performance than Raw training sets. Some significant findings are presented below:

\vspace*{2mm}
\begin{itemize} 
\itemsep=5pt
\item Data cleaning techniques (ENN, NCL, and OSS) showed low performance for datasets with Profile A.
\item In general, Random, SMOTE, and ADASYN show good performance across classifiers for datasets with Profile A.
\item There is no consistent treatment across classifiers for datasets with Profile B. Therefore, the best classification performance for datasets with high overlap and low class imbalance is obtained by selecting carefully the combination between treatment and classifier.
\item For NB (Naive Bayes) classifier the best remediation techniques for datasets with Profile B are ENN and Raw.
\item For k-NN (k-Nearest Neighbors) classifier the best remediation techniques for datasets with Profile B are OSS and NCL .
\item BDSMOTE technique takes top places associated with Profile B for LOGREG (Logistic Regression), CART (Classification and Regression Trees), and SVM-P (Support Vector Machines with Gaussian Kernel).
\item Raw (original datasets) showed low performance for SVM-G (Support Vector Machines with Gaussian Kernel) and  RBFN (Radial Basis Function Network).
\end{itemize}

\vspace*{2mm}


\begingroup
\renewcommand{\arraystretch}{0.5}
\setlength{\fboxsep}{1.5mm} 
\setlength{\tabcolsep}{0.0mm}
\begin{table}[t]
\scriptsize
    \caption{Mean ranks of the \textit{G-mean} metric based on Dataset Degradation Profile (DDP) across 49 classifiers and 9 treatments. High mean rank values tend to increase the steepness of the color gradient to green. On the the other hand, low mean rank values tend to decrease the steepness of the color gradient to red.}
    
    \label{tbl11}
    \begin{subtable}[t]{.5\textwidth}
        \caption{Profile A: high class imbalance and high disjuncts.}
        \raggedright
            \begin{tabular}{l*{10}{G}}
        &\multicolumn{9}{c}{\textbf{Treatment}} \bigskip \\ 

        \textbf{Classifier} & 
        \rotz{\rotatebox{90}{Raw}} & 
        \rotz{\rotatebox{90}{Random}} & 
        \rotz{\rotatebox{90}{SMOTE}} & 
        \rotz{\rotatebox{90}{B-SMOTE}} & 
        \rotz{\rotatebox{90}{DBSMOT}} &  
        \rotz{\rotatebox{90}{ADASYN}} & 
        \rotz{\rotatebox{90}{ENN}} & 
        \rotz{\rotatebox{90}{NCL}} & 
        \rotz{\rotatebox{90}{OSS}} \\ 
         
        NB        & 4.37	&6.00	&6.87	&7.37	&5.12	&6.37	&4.37	&3.37	&3.12 \\
        LOGREG    & 2.00	&8.00	&7.87	&6.00   &6.37	&6.37	&2.62	&4.12	&3.62  \\      
        LDA       & 3.50	&8.12	&6.75	&6.12	&6.37	&6.87	&3.62	&3.37	&3.87   \\    
        QDA       & 4.00	&7.25	&6.37	&5.62	&3.50	&7.12	&3.37	&5.00	&5.00    \\    
        C5.0      & 3.12	&6.75	&7.25	&5.25	&5.50	&7.50	&2.87	&4.00	&3.62   \\    
        CART      & 3.50    &6.87	&8.25	&4.75	&4.62	&8.00	&3.12	&4.75	&3.25    \\    
        k-NN      & 2.62	&7.75	&7.25	&5.37	&5.75	&7.00	&3.00	&4.62	&3.50     \\    
        MLN       & 2.87	&5.75	&6.37	&6.37	&4.62	&7.75	&3.25	&4.12	&4.62    \\    
        SVM-G     & 1.87	&8.12	&7.87	&6.37	&4.87	&7.75	&2.25	&4.00	&3.37    \\    
        SVM-P     & 2.50    &8.12	&7.75	&7.62	&6.12	&6.25	&2.87	&3.87	&3.50    \\    
        RF        & 3.00	&5.37	&8.00	&3.87	&6.62	&7.75	&3.62	&4.62	&3.62   \\    
        SGB       & 2.37	&7.25	&7.87	&6.00	&4.87	&7.62	&2.12	&4.37	&3.12     \\    
        ADABOOTS  & 2.62	&3.62	&8.00	&5.37	&6.75	&8.12	&4.25	&3.87	&3.37  \\    
        RBFN      & 1.87	&5.87	&8.00	&5.50	&6.50	&7.00	&3.00	&5.00	&3.50     \\

    \end{tabular}
    \end{subtable}%
   \begin{subtable}[t]{.5\textwidth}
        \raggedleft
        \caption{Profile B: high overlap and low class imbalance.}
        \begin{tabular}{l*{10}{G}}
        &\multicolumn{9}{c}{\textbf{Treatment}} \bigskip \\ 

        \textbf{Classifier} & 
        \rotz{\rotatebox{90}{Raw}} & 
        \rotz{\rotatebox{90}{Random}} & 
        \rotz{\rotatebox{90}{SMOTE}} & 
        \rotz{\rotatebox{90}{B-SMOTE}} & 
        \rotz{\rotatebox{90}{DBSMOT}} &  
        \rotz{\rotatebox{90}{ADASYN}} & 
        \rotz{\rotatebox{90}{ENN}} & 
        \rotz{\rotatebox{90}{NCL}} & 
        \rotz{\rotatebox{90}{OSS}} \\ 
         
        NB        & 6.50	&6.31	&5.18	&4.31	&3.75	&5.25	&7.18	&5.50	&5.06 \\
        LOGREG    & 3.68	&6.25	&6.00	&5.75	&6.31	&5.12	&5.18	&3.81	&4.68  \\      
        LDA       & 3.86	&7.20	&6.00	&5.93	&5.06	&5.86	&3.73	&4.33	&4.86   \\    
        QDA       & 4.58	&5.58	&4.83	&3.16	&6.00	&5.66	&6.08	&6.75	&6.16    \\    
        C5.0      & 4.25	&5.25	&5.93   &6.81	&5.43	&5.00	&5.43	&4.56	&5.50   \\    
        CART      & 4.87	&6.87	&4.93	&4.25	&6.62	&4.43	&5.18   &5.87	&6.25    \\    
        k-NN      & 3.68	&5.50	&4.68	&4.37	&5.62	&5.18	&4.25	&6.25	&6.81     \\    
        MLN       & 4.50	&5.87	&5.31	&5.62	&5.25	&5.50	&4.87	&5.31	&4.93    \\    
        SVM-G     & 2.37	&7.00	&6.31	&6.00	&5.75	&6.81	&3.43	&6.12	&4.75    \\    
        SVM-P     & 3.31	&5.81	&5.93	&4.93	&6.50	&5.93	&4.18	&6.00	&5.06   \\    
        RF        & 4.12	&6.25	&6.93	&6.00	&4.50	&7.00	&3.93	&5.12	&5.00   \\    
        SGB       & 4.18	&6.87	&6.00	&6.68	&5.87	&4.68	&4.50	&5.31	&5.43     \\    
        ADABOOTS  & 3.68	&5.81	&6.12	&5.93	&5.62	&6.31	&5.18	&5.00	&5.18  \\    
        RBFN      & 3.06	&5.18	&7.12	&4.31	&5.50	&6.43	&3.93	&5.56	&5.31     \\           

        \end{tabular}
    \end{subtable}
\end{table}

\endgroup

\newpage

\section{Conclusions and Future Work}

The performance of supervised learning algorithms is hindered by the presence of degradation problems and lack of diagnosis for training sets before building classifiers. This research describes how the introduction of diagnostic methods help minimize degradation in classification performance.

In this  research, an experimental framework was proposed to describe the complete diagnostic method, where experimental results were derived from statistical analysis over every combination of the 49 datasets, 14 classifiers, and 9 treatments (techniques for handling degradation problems). Experimental results in this study allowed to implement statistical validation procedure related to performance degradation in supervised learning. 1)  The more the class imbalance, the bigger the classifier performance degradation for metrics \textit{G-mean} and \textit{F1-score}. On the other hand, \textit{AUC} is not affected by the imbalance problem since it is an optimistic and misleading performance metric. 2) It is not true that the greater number of disjuncts, the bigger the classifier performance degradation; however, disjuncts detection allows knowing the imbalance ratio within classes. 3) It is true that the larger the overlap regions, the bigger the classifier performance degradation.

\vspace*{3mm}

In general, the SMOTE treatment takes top places in terms of increasing the classification performance, even overpassing its most recent extensions such as Borderline-SMOTE and DBSMOTE. However, SMOTE based techniques ignore the problem of small-disjuncts, more specifically subclass imbalance (imbalance within classes).

Even though data cleaning methods are designed to handling overlap problems, experiments over datasets with Profile B (high overlap and low class imbalance) showed poor classification performance. In general, results for the three data cleaning techniques used in this research (ENN, NCL, and OSS) were overpassed by sampling techniques. This can be explained by the wrong generalization of the concept of "Noisy Labels" or "mislabeling". In other words, instances located in overlap regions not necessarily have wrong labels. Thereby, these strategies of removing or label reallocation increase the problem of sparseness and reduce the classification performance. This latter result is a generalization of the Quinlan \cite{J.R.Quinlan1986} results applied to decision trees.

Results from datasets with high levels of class imbalance and disjuncts support that classification performance is significantly improved by increasing the number of synthetic Positive instances. In contrast, cleaning techniques such as ENN, NCL, and OSS showed adverse effects on classification performance.

\vspace*{3mm}

Future work is related to building a technique based on diagnostic tests and DDP, which has its primary focus of reversing the degradation effects due to class imbalance between-classes and within-classes (small-disjuncts), overlapping and noisy labels problems.

\newpage{}


\begin{table}[t]
\small
\caption{Classification algorithms, tuning parameters, and packages in \texttt{R} open source software. Abbreviation names: NB (Naive Bayes), LOGREG (Logistic Regression), LDA (Linear Discriminant Analysis), QDA (Quadratic Discriminant Analysis), C5.0, CART (Classification and Regression Trees), k-NN (k-Nearest Neighbors), MLN (Multilayer Neural Network), SVM-G (Support Vector Machines with Gaussian Kernel), SVM-P (Support Vector Machines with Polynomial Kernel), RF (Random Forest), SGB (Stochastic Gradient Boosting), ADABOOST (AdaBoost Classification Trees), and RBFN (Radial Basis Function Network).}\label{tbl12}
   \centering
   \begin{tabular}{llll}
     \toprule
     Classifier	& Function & Package & Tuning parameters\\
     \midrule
     \multirow{3}{*}{NB} & \multirow{3}{*}{nb} & \multirow{3}{*}{klaR} &
        fL (Laplace Correction)\\
        &&&usekernel (Distribution Type)\\
        &&&adjust (Bandwidth Adjustment)\\
        \midrule
        LOGREG &	glm &	R Stats &	NA\\
        \midrule
        LDA	& lda &	MASS &	NA\\
        \midrule
        QDA &	qda	& MASS &	NA\\
        \midrule
    \multirow{2}{*}{C5.0} & \multirow{2}{*}{C5.0} & \multirow{2}{*}{C5.0, plyr} &
         trials = c(1,5,10,15,20) \\
         &&&winnow = c(T,F), model = tree\\
         \midrule
        CART&	rpart&	rpart&	method = clas, cp = 0, xval = 10\\
        \midrule
        k-NN&	knn&	class&	k  (Num. Neighbors)\\
        \midrule
        
     \multirow{3}{*}{MLN} & \multirow{3}{*}{nnet} & \multirow{3}{*}{nnet} &
        size (Num. Hidden Units) \\
        &&&decay  (Weight Decay)\\
        &&&bag (Baggaing)\\
        \midrule
    \multirow{2}{*}{SVM-G} & \multirow{2}{*}{ksvm} & \multirow{2}{*}{kernlab} &
        type = C-svc,  kernel= rbfdot\\
        &&&kpar = automatic, C=1\\
        \midrule
    \multirow{2}{*}{SVM-P} & \multirow{2}{*}{ksvm} & \multirow{2}{*}{kernlab} &    
        type = C-svc,  kernel= polydot\\
        &&&kpar = automatic, C=1\\
        \midrule
        RF&	rf&	randomForest&	mtry (Num. Randomly Selected Predictors)\\
        \midrule
    \multirow{4}{*}{SGB} & \multirow{4}{*}{gbm} & \multirow{4}{*}{gbm, plyr} &
        n.trees (Num. Boosting Iterations)\\
        &&&interaction.depth (Max Tree Depth)\\
        &&&shrinkage (Shrinkage)\\
        &&&n.minobsinnode (Min. Terminal Node Size)\\
        \midrule
        ADABOOST&	adaboost&	fastAdaboost&	nIter (Num. Trees), method\\
        \midrule
        \multirow{2}{*}{RBFN} & \multirow{2}{*}{rbfDDA} & \multirow{2}{*}{RSNNS} &  
        negativeThreshold\\
        &&&(Activation Limit for Conflicting Classes)\\
     \bottomrule
   \end{tabular}
   
\end{table}

\clearpage


\begin{table}[t]
\small
\caption{Treatments for handling degradation problems and packages in \texttt{R} open source software. Abbreviation names: Raw (original set), Random (resampling), SMOTE (Synthetic Minority Oversampling Technique), B-SMOTE (Borderline-SMOTE), DBSMOTE (Density-base SMOTE), ADASYN (Adaptative Synthetic Sampling Approach for Imbalanced Learning), ENN (Edited Nearest Neighbors), NCL (Neighborhood Cleaning Rule NCL), and OSS (One-Side Selection).}\label{tbl13}
   \centering
   \begin{tabular}{llll}
     \toprule
     Name	& Function & Package & Tuning parameters\\
     \midrule
     Raw	& - & - & -\\
     \midrule
    \multirow{2}{*}{Random} & \multirow{2}{*}{-} & \multirow{2}{*}{-} &
         Over (Oversampling parameter) \\
         &&&Under (Undersampling parameter)\\
     \midrule
    \multirow{2}{*}{SMOTE} & \multirow{2}{*}{SMOTE} & \multirow{2}{*}{DMwR} & 
         perc.over, perc.under\\
         &&&k = 3\\
         &&&Over (Oversampling parameter) \\
         &&&Under (Undersampling parameter)\\
    \midrule
    \multirow{4}{*}{B-SMOTE} & \multirow{4}{*}{BLSMOTE} & \multirow{4}{*}{smotefamily} & 
     K = 3 (Num. Neighbors sampling process)\\
     &&&C = 3 (Num. Neighbors safe-level process)\\
     &&&dupSize (Num. Synthetic Positive instances)\\
     &&&method = `type2'\\
    \midrule
     DBSMOTE &	DBSMOTE	& smotefamily &	dupSize (Num. Synthetic Positive instances)\\
     \midrule
     ADASYN	& ADASYN &	smotefamily &	K = 3 (Num. Neighbors sampling process)\\
     \midrule
     ENN &	ubENN &	unbalanced & K = 3 ( Neighbors)\\
     \midrule
     NCL & ubNCL &	unbalanced & K = 3 (Num. Neighbors)\\
     \midrule
     OSS &	ubOSS &	unbalanced &	-\\
    
     \bottomrule
   \end{tabular}
   
\end{table}

Table \protect\ref{tbl14} describes main characteristics of the dataset collection, which varies in their size, number of features (Feat.), and class imbalance. First 35 datasets are from the UCI machine learning repository \cite{Dua:2019}, 7 datasets, identified with `$\ast$', from \verb|R| packages, 6 datasets from different repositories, and one new dataset, the "Simulated dataset", generated by the author. Those multiclass datasets marked with `$\dagger$' were binarized using the original dataset description. Finally, datasets identified with `$\bullet$' are subsets of the original datasets. Abbreviation names: id (dataset identification), Name (dataset name), Feat. (num. of features), Obs. (num. of observations), Mis. (num. or missing values), Dup. (num. of duplicated observations), IR trn (IR training dataset), IR test (IR testing dataset), \%Train (percentage training), \%Test (percentage testing), \%Over (percentage oversampling), \%Under (percentage undersampling), IR$\star$ (new IR trn after treatment).

\vspace*{3mm}


\begin{table}
\small
\caption{Description of datasets}\label{tbl14}
   \centering
   \begin{tabular}{clccccccccccc}
     \toprule
     id	& Name & Feat. & Obs. & Mis. & Dup. & IR trn &	IR test	&  \%Train & \%Test & \%Over & \%Under & IR$\star$\\
     \midrule
        2&EEG&14&14980&0&0&1.23&1.23&70.51&29.49&20&600&1.00\\
        3&Blocks$\dagger$&10&5473&0&45&45.44&49.63&70.15&29.85&2000&105&1.00\\
        4&ILPD$\dagger$&9&583&4&8&2.50&2.45&70.40&29.60&100&200&1.00\\
        5&Glass$\dagger$&9&214&0&1&1.88&2.60&66.20&33.80&100&180&1.11\\
        6&QSAR&41&1055&0&1&2.11&1.71&68.41&31.59&110&190&1.00\\
        7&Ozone&72&2534&687&0&13.76&12.67&71.14&28.86&1200&108&1.00\\
        8&Occupancy&5&17895&0&867&3.21&3.76&42.73&57.27&200&150&1.00\\
        9&Vertebral&6&310&0&0&2.45&1.53&69.03&30.97&200&120&1.00\\
        10&Haberman&3&306&0&7&3.02&2.13&68.56&31.44&200&150&1.00\\
        11&Spam&57&4601&0&238&1.51&1.53&69.20&30.80&50&300&1.00\\
        12&Gamma&10&19020&0&60&1.86&1.87&70.46&29.54&100&185&1.00\\
        13&Blood&4&748&0&131&2.56&3.48&62.24&37.76&200&125&1.20\\
        14&Seismic&11&2584&0&2&13.50&16.23&71.30&28.70&1000&110&1.00\\
        15&Wilt&5&4839&0&14&57.45&1.67&89.64&10.36&2000&105&1.00\\
        16&Abalone$\dagger$&7&4177&0&0&19.89&24.46&69.52&30.48&1000&110&1.00\\
        17&Audit&8&776&1&26&1.98&1.57&68.49&31.51&100&198&1.00\\
        18&Wireless$\dagger$&7&2000&0&0&2.94&3.16&70.45&29.55&100&200&1.00\\
        19&User$\dagger$&5&403&0&0&9.75&4.58&64.02&35.98&900&105&1.07\\
        20&Shuttle$\dagger\bullet$&9&17500&0&0&4.00&3.84&71.43&28.57&300&133&1.00\\
        21&Image$\dagger$&18&2310&0&120&1.27&1.43&69.41&30.59&30&420&1.00\\
        22&Happiness&6&143&0&9&1.08&0.72&58.96&41.04&20&530&1.22\\
        23&Skin&3&12500&0&3723&2.38&4.00&57.97&42.03&100&200&1.00\\
        24&Seeds&7&210&0&0&1.82&2.45&67.14&32.86&100&182&1.11\\
        25&Musk1&166&476&0&0&1.40&1.08&71.22&28.78&50&280&1.06\\
        26&CTG$\dagger$&20&2129&3&8&3.58&3.40&70.30&29.70&200&170&1.13\\
        27&Sonar&60&208&0&0&1.18&1.06&69.23&30.77&20&590&1.00\\
        28&Forest$\dagger$&27&523&0&0&4.35&6.07&37.86&62.14&400&109&1.16\\
        29&HTRU2&8&17898&0&0&9.90&9.97&70.45&29.55&800&110&1.00\\
        30&Adult$\oplus$&6&26281&0&24&3.13&3.23&37.99&62.01&200&150&1.00\\
        31&Sports&59&1000&0&3&1.86&1.52&67.60&32.40&100&185&1.08\\
        32&Banknote&4&1372&0&11&1.21&1.28&69.29&30.71&20&600&1.00\\
        33&Electrical&13&10000&0&0&1.78&1.73&70.43&29.57&100&177&1.13\\
        34&Wine$\dagger$&13&178&0&0&2.94&2.20&73.03&26.97&200&146&1.00\\
        35&Breast&9&699&0&153&1.03&0.49&66.79&33.21&10&1034&1.14\\
        36&nanoHUB&18&14110&0&142&39.63&38.02&70.39&29.61&2000&105&1.00\\
        37&Simulated&2&700&0&0&4.70&5.18&70.00&30.00&300&130&1.00\\
        38&Weather$\ast$&17&266&0&0&4.52&4.55&68.64&31.36&400&113&1.11\\
        39&Pima1$\ast$&7&532&0&0&2.11&1.76&71.99&28.01&100&200&1.00\\
        40&Ionosphere$\ast$&32&351&0&1&1.76&1.96&79.71&20.29&76&233&1.00\\
        41&Pima2$\ast$&8&768&0&0&2.01&1.58&69.40&30.60&200&150&1.00\\
        42&Ringnorm&20&7400&0&0&1.02&1.02&70.31&29.69&2&5100&1.00\\
        43&College$\ast$&17&777&0&0&3.05&2.03&69.24&30.76&200&152&1.00\\
        44&Iris$\ast\dagger$&4&150&0&0&1.74&2.56&62.00&38.00&76&228&1.00\\
        45&MDRR$\ast$&342&528&0&2&1.36&1.16&71.67&28.33&40&339&1.00\\
        46&Mammo&6&11183&0&2343&28.56&46.58&62.86&37.14&1000&110&1.00\\
        47&Phoneme&5&5404&0&32&2.39&2.48&70.16&29.84&100&200&1.00\\
        48&Fraud$\bullet$&30&10492&0&13&21.94&18.69&70.49&29.51&1000&110&1.00\\
        49&SDSS$\dagger\bullet$&15&10000&0&0&11.28&9.74&69.39&30.61&1000&110&1.00\\
     \bottomrule
   \end{tabular}
   
\end{table}

\clearpage


\makeatletter 
\g@addto@macro{\@algocf@init}{\SetKwInOut{paramete}{paramete}} 
\makeatother
\begin{algorithm}[H]
\small
    \SetEndCharOfAlgoLine{}
    
    \newcommand\mycommfont[1]{\footnotesize\ttfamily\textcolor{blue}{#1}}
    \SetCommentSty{mycommfont}

    \SetKwInOut{Input}{Input}
    \SetKwInOut{Output}{Output}
 
    \Input{%
    \begin{tabular}[t]{ll}
      $S$ & is the training set, $X = \{x_1,x_2,...,x_{i+1}, x_{i+2},...x_M\}$ are the instances of $S$ \\
      $Label$ & is the class membership vector of  $X_i$. Positive ($P$) or Negative ($N$)\\
      $k$ & is the number of k-Nearest Neighbors \\
      $C.min$ & is the minimum num. of instances per component (subclass)\\
      $G.max$& is the maximum num. of mixture components for which the BIC is calculated\\
      $sp.th$ & is the separation index threshold\\
      $dist$ & is the distance measure to be use. 1 = Mahalanobis, 2 = Manhattan, and 3 = Euclidean\\
    \end{tabular}
  }
  
   \Output{%
    \begin{tabular}[t]{ll}
        $S_1$ & if $C.min$ is not covered, $S_1=$ \textit{Subclass membership} for outlier analysis\\
        $S_2$\hspace{6mm}  & if $C.min$ is covered, $S_2= S_1 +$ \textit{Diagnostic report}\\
    \end{tabular}
  }
  
  \BlankLine
      
  \textbf{Diagnosis}$(S,k,C.min,G.max,sp.th,dist)$\\
  
   
   \tcc{Find subclasses by Gaussian Mixture Models}
   \ForEach{Class $\in$ $S$}{ 
     \textit{Best Model} = $\argmax_{x \in Class}$ BIC{$\left(\mathbf{X};\theta\right)$}, \text{see Eq. (5)}\\ 
     Let $S_1 = \{$\textit{Subclass membership} $\cup\ S\}$\;
     $P_g = \{p_1,...,p_{K_P}\}$, set of subclasses from P; $K_P = $ Num. of subclasses in $P$\;
     $N_j = \{n_1,...,n_{K_N}\}$, set of subclasses from N; $K_N = $ Num. of subclasses in $N$\;
     $L_P = Length(P_g)$, vector with the length of each subclass P\;
     $L_N = Length(N_j)$, vector with the length of each subclass N\;
    }
   \eIf{\textbf{any} $(L_p,L_N)$  < C.min}{
            Return $S_1$\;
            }{
            \tcc{Calculate IR between subclasses}
                    $Upper =$ \textit{IR}$(Sets)$, upper triangular matrix\;
                    $Diag =$ {$L_P$ $\cup$ $L_N$}, diagonal of the triangular matrix\;

             \tcc{Find the optimal projection between subclasses} 
             \eIf{Projection is TRUE}{
                   $Lower =  J^*(S_1)$, lower triangular matrix, see Eq. (7)\;
                 }{
                   $Lower =$  IR$(S_1)$, lower triangular matrix, see Eq. (7)\;
                  }
                  \textit{IRO} $= \{Upper \cup Lower \cup Diag\}$, Imbalance Ratio and Overlap matrix\;
                  \tcc{Noisy label analysis}
            
                 \ForEach{$X_i \in S$}{
                    Find \textit{class.neighbors}, class with largest num. of instances among the k-Nearest Neighbors\;
                    \eIf{$Label_i = class.neighbors$}{
                      $x_i$ has the correct label\;
                    }{
                      $x_i$ is noise\;
                        \eIf{$x_i$ is located between two subclasses with $J^* \geq sp.th$}{
                          $x_i \in$ \textit{NL}, Noisy Label set\;
                        }{
                          $x_i \in$ \textit{NO},  Noise Overlap set\;
                        }
                    }
                 }
                \tcc{Dispersion analysis: DA}
                \ForEach{Class $\in S$}{
                    Estimate $mean$ and $std$ to the $median$ per subclass by using $dist$\;
                    Execute the test for homogeneity of multivariate dispersion
                }
        \textit{Diagnostic report} = \textit{IRO}, \textit{NL}, \textit{NO}, and \textit{DA}\;
        Let $S_2 = S_1 +$ \textit{Diagnostic report}\;
        Return $S_2$\;
    }
\caption{Diagnostic tests}\label{alg1}
\end{algorithm}

\newpage

\appendix

\section{Gaussian Mixture Models and Separation Index}

\textbf{Gaussian Mixture Models (GMM):} the Multivariate Gaussian distribution can be defined over a $D$-dimensional vector $x$ of contiguous variables, which is given by

\begin{equation}\label{eq1} 
\mathcal{N}\left(x\middle|\mu,\Sigma\right)=\frac{1}{\left(2\pi\right)^\frac{D}{2}}\frac{1}{\left|\Sigma\right|^\frac{1}{2}}\exp{\left\{-\frac{1}{2}\left(x-\mu\right)^T\Sigma^{-1}\left(x-\mu\right)\right\}}
\end{equation}

where $\mu$ is the mean vector of dimension $D$, $\Sigma$ is the covariance matrix of dimension $D$$\times$$D$, and $|\Sigma|$ denotes the determinant of $\Sigma$.
Then, GMM can be written as a linear superposition of  $K$ Multivariate Gaussian Distributions \cite{Pearson2006,titterington_85} in the form

\begin{equation}\label{eq2} 
p\left(x|\pi,\mu,\Sigma\right)=\sum_{k=1}^{K}\pi_k\mathcal{N}\left(x\middle|\mu_k,\Sigma_k\right)
\end{equation}

Where $\pi_k=\ \left\{\pi_1,\ldots\pi_K\right\}$ is the mixing coefficient vector (probability vector of membership) and $\sum_{k=1}^{K}{\pi_k=1}$ in order to be valid probabilities. Moreover,  $\mu_k=\ \left\{\mu_1,\ldots\mu_K\right\}$ and $\Sigma_k=\ \left\{\Sigma_1,\ldots\Sigma_K\right\}$.

For a given training set  $\mathbf{X}=\ \left\{x_1,\ \ldots,\ x_N\right\}$, $\hat{\mathcal{L}}$ is the Maximum Likelihood Estimator (LME) and its log-likelihood function is expressed by \verb|Equation| \ref{eq4} , which estimates the probability of membership of $K$ components. The EM algorithm is used to maximize this function and estimate the mixture parameters \cite{A.P.DempsterN.M.Laird1977}.

\begin{equation}\label{eq3} 
\hat{\mathcal{L}}{\left(\mathbf{X};\theta\right)}=p\left(\mathbf{X}|\pi,\mu,\mathrm{\Sigma}\right)\
\end{equation}

\begin{equation}\label{eq4} 
log{p\left(\mathbf{X}|\pi,\mu,\mathrm{\Sigma}\right)}{=\sum_{n=1}^{N}log{\left\{\sum_{k=1}^{K}{\pi_k\mathcal{N}\left(x\middle|\mu_k,\mathrm{\Sigma}_k\right)}\right\}}}\
\end{equation}

Determining the optimal number of components $K$ continues to be a problem for empirical cluster techniques such as K-means. However, probabilistic approaches based on mixture models can solve this problem using a penalty function. If the number of components $K$ is increased in the \verb|Equation| \ref{eq4}, it results in an increase in the dimensionality of the model, causing a monotonous increase in its likelihood. This is particularly a problem because it is not useful to obtain as many components as instances. Therefore, the best GMM is the one that maximizes the Bayesian Information Criterion (BIC) \cite{Schwarz1978,Fraley1998a}, which seeks to balance the increase in likelihood and the complexity of the model by introducing a penalty term for each parameter. BIC is defined by

\begin{equation}\label{eq5} 
BIC{\left(\mathbf{X};\theta\right)}=log{p\left(\mathbf{X}|\pi,\mu,\mathrm{\Sigma}\right)}-\frac{1}{2}\eta\left(\theta\right)log{\left(N\right)}\ 
\end{equation}

where $\eta(\theta)$ is the number of free parameters in the model that represents the complexity of the model.
GMM build ellipsoidal subclasses (clusters), centered at the means $\mu_k$ and with covariance $\mathrm{\Sigma}_k$. The covariance matrix can be determined by geometric characteristics. For instance, Banfield et al. \cite{Banfield1993} develop a covariance matrix model in terms of its eigenvalues decomposition in the term form

\begin{equation}\label{eq6} 
\mathrm{\Sigma}_k=\lambda_kD_k{A_kD}_k^T\ 
\end{equation}

where $D_k$ is the orthogonal matrix of eigenvectors and defines the orientation,  $A_k$ is a diagonal matrix whose elements are proportional to the eigenvalues of $\mathrm{\Sigma}_k$ and determines the shape of the density contours,  and $\lambda_k$ is a scalar that defines the volume of the corresponding ellipsoid. This covariance model is the generalization of several earlier proposals based on GMM. Table \protect\ref{tbl15} is taken for the \verb|mclust| package \cite{Scrucca2016,Fraley1998} in \verb|R| open source software, which shows names, models, and geometric interpretation of the covariances model implemented for this research.

\vspace{4mm}


\begin{table}[t]
\small
   \centering
   \caption{Parametrization of the covariance matrix for Gaussian models. Source \cite{Scrucca2016}}
   \label{tbl15}
   \begin{tabular}[t]{cccccc}
     \toprule
    Model   &$\mathrm{\Sigma}_k$         &Distribution  &Volume      &Shape      &Orientation \\
     \midrule
      EII	&$\lambda I$                &Spherical	   &Equal	    &Equal	    &$-$  \\
      VII	&$\lambda_k I$              &Spherical	   &Variable	&Equal	    &$-$  \\
      EEI	&$\lambda A$                &Diagonal	   &Equal	    &Equal	    &Coordinate axes \\
      VEI	&$\lambda_k A$              &Diagonal	   &Variable	&Equal	    &Coordinate axes  \\
      EVI	&$\lambda A_k$              &Diagonal	   &Equal	    &Variable	&Coordinate axes \\
      VVI	&$\lambda_k A_k$            &Diagonal	   &Variable	&Variable	&Coordinate axes  \\
      EEE	&$\lambda DAD^T$            &Ellipsoidal   &Equal	    &Equal	    &Equal  \\
      EVE	&$\lambda DA_kD^T$          &Ellipsoidal   &Equal	    &Variable	&Equal  \\
      VEE	&$\lambda_k DAD^T$          &Ellipsoidal   &Variable	&Equal	    &Equal  \\
      VVE	&$\lambda_k DA_kD^T$        &Ellipsoidal   &Variable	&Variable	&Equal  \\
      EEV	&$\lambda D_k{AD}_k^T$      &Ellipsoidal   &Equal	    &Equal	    &Variable  \\
      VEV	&$\lambda_k D_k{AD}_k^T$    &Ellipsoidal   &Variable	&Equal  	&Variable \\
      EVV	&$\lambda D_k{A_kD}_k^T$    &Ellipsoidal   &Equal	    &Variable	&Variable  \\
      VVV	&$\lambda_kD_k{A_kD}_k^T$   &Ellipsoidal   &Variable	&Variable	&Variable \\
  
     \bottomrule
   \end{tabular}
\end{table}


\newpage{}

\textbf{Separation index:} this index ($J^\ast$) was proposed by Qiu et al. \cite{Qiu2006,Qiu2006a}. This index measures the magnitude of the gap between pairs of subclasses. It has a value between $-$1 and +1, where negative values indicate subclasses are overlapped, zero means subclasses are touching, and positive values indicate subclasses are separated. Initially, it is necessary to find the optimal projection in one-dimension space in which two subclasses have the maximum separation. 
The initial projection is selected between two possible methods:  $\left(\mathrm{\Sigma}_1+\mathrm{\Sigma}_2\right)^{-1}\left(\mu_2-\mu_1\right)$ and $\left(\mu_2-\mu_1\right)$. Then, the Newton-Raphson method is used to search the optimal projection. Therefore, it requires that both covariances matrices from subclasses must be positive definite. This assumption is covered because all subclasses follow a multivariate Gaussian distribution. In a nutshell, the covariance matrix must be a symmetric positive definite matrix in order to the \verb|Equation| \ref{eq7} makes sense.

Let $J_{12}^\ast$ be the optimal separation index between subclasses 1 and 2. Then, $J_{12}^\ast=J_{12}(a^\ast)$ , where $a^\ast$ is the optimal projection direction witch maximizes $J_{12}$. Finally, the Separation index $J^\ast$ is estimated by using \verb|Equation| \ref{eq7}.

\begin{equation}\label{eq7} 
{J(a}^\ast)=\frac{a^T\left(\mu_2-\mu_1\right)-q_\frac{\alpha}{2}\left(\sqrt{a^T\mathrm{\Sigma}_1a}+\sqrt{a^T\mathrm{\Sigma}_2a}\right)}{a^T\left(\mu_2-\mu_1\right)+q_\frac{\alpha}{2}\left(\sqrt{a^T\mathrm{\Sigma}_1a}+\sqrt{a^T\mathrm{\Sigma}_2a}\right)}
\end{equation}

where $\mu_1$, $\mu_2$, $\mathrm{\Sigma}_1$, and $\mathrm{\Sigma}_2$ are the probabilistic parameters of the two subclasses, $\alpha\ \in(0,05)$ is a tuning parameter indicating the percentage of data in the extremes to downweigh, $q_{\alpha/2}=\ z_{\alpha/2}$, where $z_{\alpha/2}$ is the upper $\alpha/2$ quantile of the univariate standard normal distribution.

\section{Performance Metrics}

The confusion matrix is used to describe the performance of a classification algorithm on a set of test data for which the true classes are known. Table \protect\ref{tbl16} shows the confusion matrix elements for a binary classification problem. Traditionally, the overall accuracy ($Acc$) is the most important metric to evaluate the performance of a classifier, which can be calculated using \verb|Equation| \ref{eq8}.

\vspace{4mm}

\newpage{}

\begin{table}[t]
\small
\caption{Confusion matrix for a binary classification task}\label{tbl16}
   \centering
   \begin{tabular}{llcc}
     \toprule
     &  & \multicolumn{2}{c}{\textbf{Prediction}}\\
     &  & Positive (P) & Negative (N) \\
     \midrule
     \multirow{2}{*}{\textbf{Observations}} & 
        Positive (P) & TP & FN\\
        &Negative (N) & FP & TN\\
     \bottomrule
   \end{tabular}
\end{table}
\noindent\begin{minipage}{.5\linewidth}
\begin{equation}\label{eq8}
  Acc=\frac{TP+TN}{TP+TN+FP+FN}\ 
\end{equation}
\end{minipage}%
\begin{minipage}{.5\linewidth}
\begin{equation}\label{eq9} 
  Sensitivity= {Acc}^+ = \frac{TP}{TP+FN}
\end{equation}
\end{minipage}
\vspace*{3mm}

\noindent\begin{minipage}{.5\linewidth}
\begin{equation}\label{eq10}
  Precision=\ \frac{TP}{TP+FP}
\end{equation}
\end{minipage}%
\begin{minipage}{.5\linewidth}
\begin{equation}\label{eq11} 
  Specificity= {Acc}^- = \frac{TN}{TN+FP}
\end{equation}
\end{minipage}
\vspace*{3mm}

Sometimes the overall accuracy can be misleading \cite{Provost1998} and models with a lower overall accuracy could have better predictive power. For instance, in case of imbalance data a classifier can reach high accuracy levels for the Negative class; however, the accuracy for the Positive class is low. Therefore, the overall accuracy has a bias toward the Negative class in imbalance data. To solve this situation, more specific metrics are used such as \textit{sensitivity} (\textit{recall}, \textit{true positive rate}, or ${Acc}^+$), \textit{specificity} (\textit{true negative rate} or ${Acc}^-$) and \textit{precision} (\textit{positive predictive value}).\

The Receiver Operating Characteristic (\textit{ROC}) curve is a well-accepted technique for summarizing classifier performance \cite{Swets1988}. The area under the \textit{ROC} curve (\textit{AUC}) is a measure of how well a classifier can distinguish between two classes \cite{Bradley1997,Duda2000}.

\begin{equation}\label{eq12} 
AUC=\frac{1}{P\times N}\sum_{i=1}^{P}\sum_{j=1}^{N}1_{p_i>p_j}
\end{equation}

Where $i$ runs over all Positive instances with true class 1, and $j$ runs over all Negative instances with true class 0; $p_i$ and $p_j$ denote the probability score assigned by the classifier to instance $i$ and $j$, respectively. 1 is the indicator function: it outputs 1 if, and only if,  the condition $(p_i>p_j)$ is satisfied. In simple words, the \textit{ROC} curve is generated by plotting the "\textit{True Positive Rate}" (\textit{TPR}) against the "\textit{False Positive Rate}" (\textit{FPR}). However, This \textit{AUC} metric may provide an overly optimistic performance evaluation \cite{Davis2006}. Thereby, performance evaluation metrics such as \textit{Precision-Recall} curves, \textit{F1-score} (\textit{F-measure}), the \textit{geometric mean} (\textit{G-mean} or \textit{G-measure}), and the \textit{correlation coefficient} $\phi$ are considerate more convenient in the class imbalance context.

\vspace{4mm}

\noindent\begin{minipage}{.5\linewidth}
\begin{equation}\label{eq13}
  TPR=\ \frac{TP}{TP+FN}
\end{equation}
\end{minipage}%
\begin{minipage}{.5\linewidth}
\begin{equation}\label{eq14} 
  FPR=\ \frac{FP}{FP+TN}\ 
\end{equation}
\end{minipage}
\vspace*{2mm}

In Statistics, \textit{F1-score} (also \textit{F-score} or \textit{F-measure}) uses the \textit{Precision} and \textit{Sensitivity} providing a single accuracy measurement for a classifier \cite{He2009}. Formally, it is the harmonic average of the \textit{Precision} and \textit{Recall}, which return a value between 0 and 1. Where 1 indicates perfect \textit{Precision} and \textit{Sensitivity}.\

\begin{equation}\label{eq15} 
F1-score=\ \ 2\ \times\ \frac{Precision\ \times\ Sensitivity}{Precision+Sensitivity}
\end{equation}

The \textit{Geometric mean score} (\textit{G-mean} or \textit{G-measure}) is the geometric mean of \textit{Specificity} and \textit{Sensitivity}, which return a value between 0 and 1 \cite{Drummond2003,Japkowicz2013}.\

\begin{equation}\label{eq16} 
G-mean=\sqrt{Sensitivity\ \times\ Specificity} = \sqrt{{Acc}^+\times\ {Acc}^-}
\end{equation}

In Statistics, the phi coefficient (also phi correlation coefficient or mean square contingency coefficient) measure of the degree of association between two binary variables, which are considered positively associated if most of the data falls along the diagonal cells of the confusion matrix. Phi coefficient return a value between $-$1 and 1, where values between 0.7 to 1 indicate strong association.

\begin{equation}\label{eq18} 
\phi=\frac{TP\ \times\ TN-FP\ \times\ FN}{\sqrt{\left(TP+FP\right)\times\left(FN+TN\right)\times\left(TP+FN\right)\times\left(FP+TN\right)}}\ 
\end{equation}

\newpage
\setlength{\bibsep}{2pt plus 0.3ex}
\renewcommand*{\bibfont}{\small}
\setstretch{0.2}

\bibliographystyle{unsrt} 

\bibliography{Ref_Diagnostic_v7}

\end{document}